%% file: auxo-socc23.tex
 \documentclass[sigconf,10pt]{acmart}

\usepackage{url, cite}
\usepackage{xspace} 
\usepackage{microtype}
\usepackage{graphicx}
\usepackage{subcaption}
\usepackage{booktabs} 
\usepackage{multirow}

\usepackage{xspace}
\usepackage{tabularx}
\usepackage{tikz}

\usepackage{subcaption}
\usepackage{url, cite}
 \usepackage{verbatim}
\usepackage{multirow}
\usepackage[makeroom]{cancel}
\usepackage{dsfont}
\usepackage{amsmath}

\usepackage{tikz}
\usepackage{tabularx}
\usepackage{color, colortbl}
\usepackage{hhline}

\usepackage[linesnumbered,ruled,noend,scleft]{algorithm2e}
\usepackage{float} 
\usepackage{algorithmicx}

\definecolor{cardinal}{rgb}{0.77, 0.12, 0.23}

\SetCommentSty{mycommfont}
\newcommand{\algcomment}[1]{{\footnotesize\Comment{#1}}}

\def\name{{Auxo}\xspace}

\definecolor{egyptianblue}{rgb}{0.06, 0.2, 0.65}
\definecolor{grey}{rgb}{0.33, 0.33, 0.33}
\definecolor{yellow}{rgb}{1.0, 0.88, 0.21}

\def\ie{{i.e.\xspace}}
\def\eg{{e.g.\xspace}}

\newcommand*{\circled}[1]{\lower.7ex\hbox{\tikz\draw (0pt, 0pt)%
    circle (.5em) node {\makebox[1em][c]{\small #1}};}}

\usepackage{hyperref}
\hypersetup{
  colorlinks=true,      
  linkcolor=blue,       
  citecolor=magenta,    
  filecolor=cyan,       
  urlcolor=red          
}

\graphicspath{{figures/}}

\copyrightyear{2023}
\acmYear{2023}
\setcopyright{acmlicensed}
\acmConference[SoCC '23]{ACM Symposium on Cloud Computing}{October
30--November 1, 2023}{Santa Cruz, CA, USA}
\acmBooktitle{ACM Symposium on Cloud Computing (SoCC '23), October
30--November 1, 2023, Santa Cruz, CA, USA}
\acmPrice{15.00}
\acmDOI{10.1145/3620678.3624651}
\acmISBN{979-8-4007-0387-4/23/11}

\begin{document}

\title{\name: Efficient Federated Learning via Scalable Client Clustering}

\author{Jiachen Liu}
\affiliation{%
  \institution{University of Michigan}
  \country{}
}
\email{amberljc@umich.edu}

\author{Fan Lai}
\affiliation{%
  \institution{University of Illinois Urbana-Champaign}
  \country{}
}
\email{fanlai@illinois.edu}

\author{Yinwei Dai}
\affiliation{%
  \institution{Princeton University}
  \country{}
}
\email{yinweid@cs.princeton.edu}

\author{Aditya Akella}
\affiliation{%
  \institution{University of Texas at Austin}
  \country{}
}
\email{akella@cs.utexas.edu}

\author{Harsha V. Madhyastha}
\affiliation{%
  \institution{University of Southern California}
  \country{}
}
\email{madhyast@usc.edu}

\author{Mosharaf Chowdhury}
\affiliation{%
  \institution{University of Michigan}
  \country{}
}
\email{mosharaf@umich.edu}




\begin{abstract}

\input{pages/abstract.tex}

\end{abstract} 

\maketitle

\renewcommand{\shortauthors}{J. Liu,
F. Lai,
Y. Dai,
A. Akella,
H. V. Madhyastha,
M. Chowdhury}

\input{pages/intro.tex}

\input{pages/motivation.tex}

\input{pages/overview.tex}

\input{pages/algorithm.tex}

\input{pages/sys.tex}

\input{pages/impl.tex}

\input{pages/eval.tex}
\input{pages/related.tex}

\input{pages/conclusion.tex}

\input{pages/ack.tex}

\label{EndOfPaper}

\bibliographystyle{ACM-Reference-Format}
\bibliography{auxo}

\clearpage

\appendix
\input{pages/appendix.tex}
\end{document}

%% file: pages/abstract.tex
Federated learning (FL) is an emerging machine learning (ML) paradigm that enables heterogeneous edge devices to collaboratively train ML models without revealing their raw data to a logically centralized server. 
However, beyond the heterogeneous device capacity, FL participants often exhibit differences in their data distributions, which are not independent and identically distributed (Non-IID).
Many existing works present point solutions to address issues like slow convergence, low final accuracy, and bias in FL, all stemming from client heterogeneity.

In this paper, we explore an additional layer of complexity to mitigate such heterogeneity by grouping clients with statistically similar data distributions (\emph{cohorts}). 
We propose \name to gradually identify such cohorts in large-scale, low-availability, and resource-constrained FL populations.
\name then adaptively determines how to train cohort-specific models in order to achieve better model performance and ensure resource efficiency.  
Our extensive evaluations show that, by identifying cohorts with smaller heterogeneity and performing efficient cohort-based training, \name boosts various existing FL solutions in terms of final accuracy (2.1\%--8.2\%), convergence time (up to 2.2$\times$), and model bias (4.8\% - 53.8\%).

%% file: pages/intro.tex
\section{Introduction}
\label{sec:intro}
 
Federated learning (FL) enables distributed clients to collaboratively train an ML model without centralizing their local data to the cloud.
It circumvents the systematic privacy risk and cost of data transfers in centrally collecting user data. Hence, FL is increasingly being adopted by many popular applications, such as Google's Gboard~\citep{gboard}, Apple's Siri~\citep{siri}, NVIDIA's medical platform~\citep{nvidia}, Meta's Ads recommendation~\citep{meta}, and WeBank risk prediction~\citep{webank}.

Federated Learning (FL) typically involves a substantial number of clients, ranging from hundreds to millions, and the training process can span days or even weeks~\citep{google-query-sugg}. 
Given the limited availability and resource constraints of client devices, only a fraction of clients contribute to each round of training in practice.
Therefore, it is essential to reduce the training time while accommodating these practical constraints.
However, FL encounters unique challenges stemming from statistical heterogeneity among user data, which contributes significantly to extended training time and suboptimal model performance~\citep{convergence1,convergence2, personalization-cluster, multi-personal}. 
Several studies that try to mitigate the effect of statistical heterogeneity, such as FedYoGi~\citep{yogi}, q-FedAvg~\citep{qfedavg}, FTFA~\citep{fine-tune}, have shown that their convergence speed depends on the degree of heterogeneity, both theoretically and empirically (\S\ref{sec:flchallenge}).

 We explore the possibility of mitigating this issue at its core by grouping clients with similar data distributions, known as \emph{cohorts}~\citep{generalization} (\S\ref{sec:room-improv}). 
If a population has $K$ cohorts, training $K$ separate models -- one for each cohort with lower statistical heterogeneity -- can boost the performance of many existing FL algorithms that are complementary to ours and focus on convergence~\citep{yogi, prox, oort}, fairness optimization~\citep{qfedavg}, communication efficiency~\citep{marco, fetchsgd,pisces}, etc.

Although recent works attempted to identify cohorts and train separate models for them~\citep{hc+fl,ifca,multi-personal, icfl}, they are not applicable to real-world FL deployments. 
This is because unlike easy-to-deploy solutions such as FedAvg and FedYoGi~\citep{yogi, communication-eff}, clustering clients at scale and in the wild poses unique challenges (\S\ref{sec:challenge}). 
Existing solutions often ignore the scale and sparsity of the device participation.
They also ignore the constraints on availability and capacity of end-user devices, which calls for low-overhead algorithms.

We propose \textbf{\name} to enable 
1) scalable cohort identification to reduce intra-cohort heterogeneity in large-scale and limited-availability FL scenarios; and 
2) efficient cohort-based training to facilitate most FL optimizations, such as faster training completion and better model accuracy, without additional resource requirements.
\name addresses the following challenges toward practical FL deployment (\S\ref{sec:algo}). 
First, unlike existing clustering strategies which require exhaustive passes through all clients~\citep{cfl}, on-demand device availability~\citep{hc+fl}, or additional on-device training for every participant~\citep{ifca,flexcfl}, \name introduces a more flexible client clustering solution. It allows sporadic client availability, respects client resource constraints, and maintains client privacy. 
\name can progressively identify cohorts and scalably cluster clients based on their gradients in spite of the absence of anchored gradients for straightforward comparison. 
Second, unlike expensive and ad-hoc hyper-parameter tuning stages used in existing solutions, \name progressively generates the appropriate number of cohorts and identifies suitable timings to create them. 
Thus, \name maximizes the use of limited client resources to enhance training speed and model performance.\footnote{We refer to the number of participants that contribute to a round of FL training as training resource throughout this paper.}
Finally, we design a scalable system to support efficient cohort clustering and training at scale while being robust to uncertainties (\eg, failure tolerance and unfavorable settings) at scale (\S\ref{sec:system}).

We have implemented (\S\ref{sec:impl}) and evaluated (\S\ref{sec:eval}) \name on a wide variety of real-world FL datasets, tasks, and algorithms at scale. 
Compared to existing solutions, \name improves the performance for various FL algorithms, such as better model accuracy (2.1\%-8.2\%) and convergence speed (up to 2.2$\times$) and smaller bias of model accuracy (4.8\% - 53.8\%).

Overall, we make the following contributions in this paper:
\begin{enumerate}
    \item We propose a systematic clustering mechanism to identify cohorts for the practical large-scale, low-availability and resource-constrained FL setting.
    
    \item We identify a sweet spot for jointly optimizing model convergence and training cost, and provide analytical insights to ensure good model performance.
     
    \item We implement and evaluate \name at scale, showing large improvements in final accuracy, convergence time, and model fairness over the state-of-the-art.
    \name is open-source and available on GitHub.\footnote{https://github.com/SymbioticLab/FedScale/tree/master/examples/auxo}

\end{enumerate}

%% file: pages/motivation.tex
\section{BACKGROUND AND MOTIVATION}%
\label{sec:motivation}
 
We start with a brief introduction of federated learning (\S\ref{sec:back}),
followed by the challenges it faces in real-world settings (\S\ref{sec:flchallenge} ) .
Next, we describe some opportunities to improve FL that motivates our work (\S\ref{sec:room-improv}).
Finally, we explain the limitations of related works that motivate our algorithm and system design  (\S\ref{sec:challenge}).

\subsection{Federated Learning}
\label{sec:back}
\begin{figure}[t!]
   \centering
   \includegraphics[trim=0 130 0 0,clip,scale=0.88]{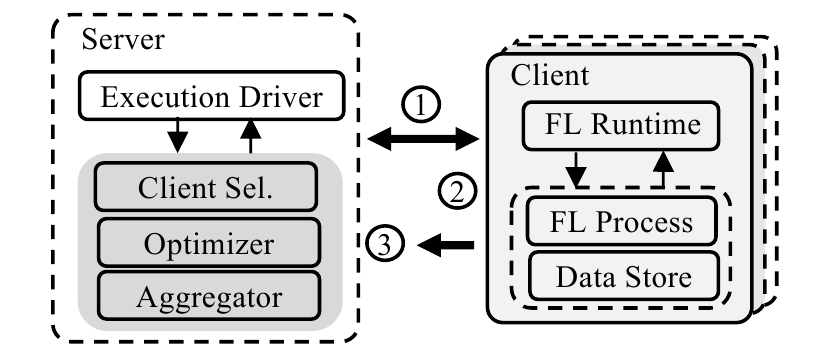} 
     \caption{Traditional FL overview. The server first selects from available clients and sends out model weights. Clients train the updated model on their local dataset. After training is finished, clients report their model gradient to the server.}
   \label{fig:orig}
   \vspace{-0.4cm}
   \end{figure}
 
A typical cross-device FL system consists of two primary components (Figure~\ref{fig:orig}):
A logically centralized cloud \emph{server} that maintains a single global model and many distributed \emph{clients} with private local data. 
The overall lifecycle of an FL training round can be divided into three broad stages.
\begin{enumerate}
  \item[\circled{1}] \textbf{Selection stage:} Clients check in with the server continuously to announce their availability for FL computation.
    The server selects a number of \emph{participants} for that round based on its client selection strategy.
 
  \item[\circled{2}] \textbf{Execution stage:} The selected participants download the current model from the server and perform server-specified computation on their local data. 
  
  \item[\circled{3}] \textbf{Aggregation stage:} Participants that successfully complete the execution stage send model updates back to the server.
    The server aggregates the updates to finalize an updated model for the next round.
\end{enumerate}

\subsection{Heterogeneity Challenges in FL} 
\label{sec:flchallenge}
 
Unlike centralized ML, FL faces unique challenges in terms of statistical and system heterogeneity.
The former refers to the varying data volumes and difference of data distribution across clients, which hinders model convergence; the latter refers to variations in system characteristics among participants' devices, which results in large differences in training performance.
Increasing heterogeneity in either dimension leads to poor performance.

\begin{figure}[t!]
  \centering
     \begin{subfigure}{1.6in}
  \centering
   \includegraphics[scale=0.3]{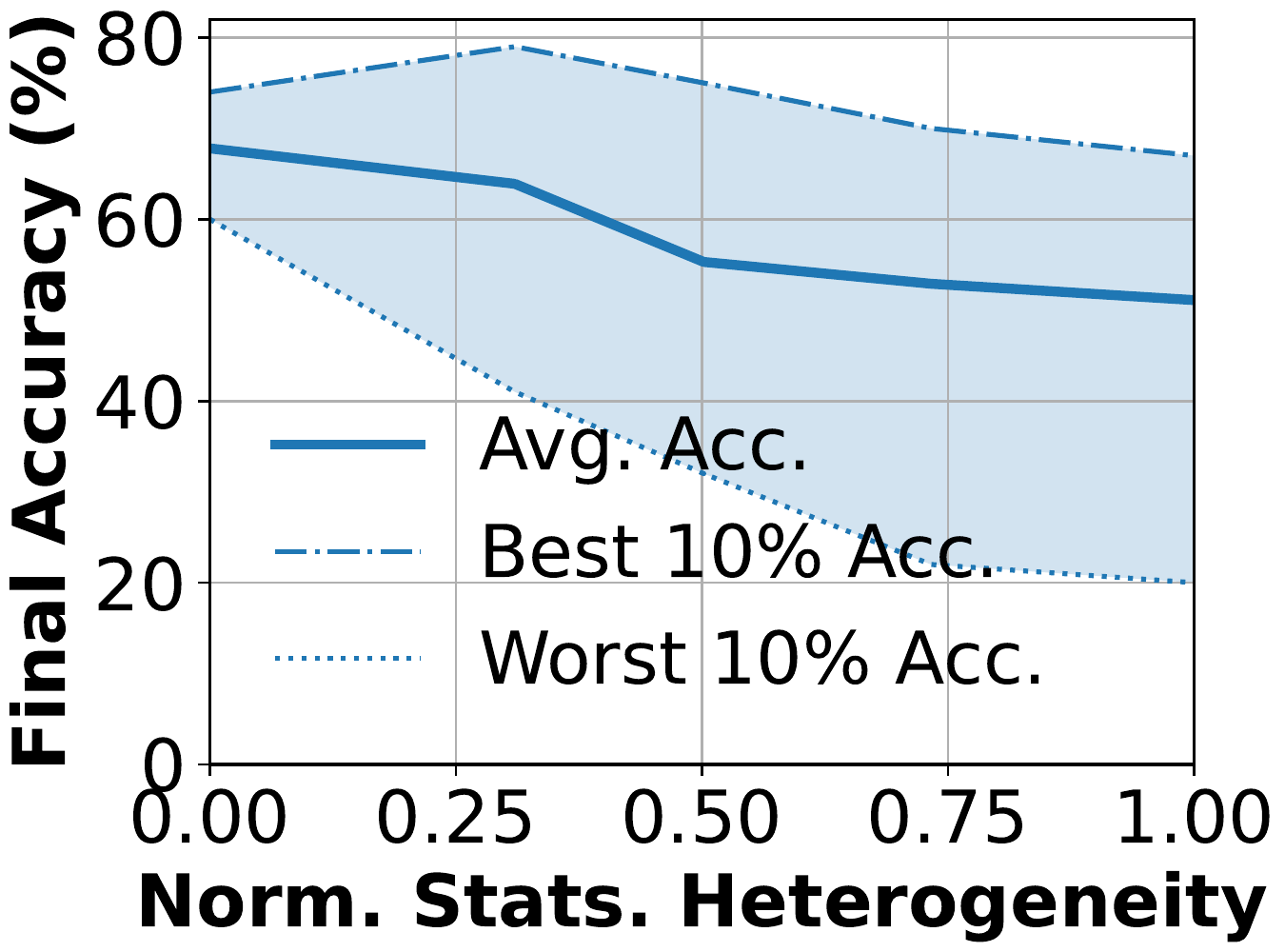} 
     \caption{Statistical heterogeneity. }
   \label{fig:heter}
  \end{subfigure}
     \begin{subfigure}{1.6in}  
   \centering 
   \includegraphics[scale=0.3]{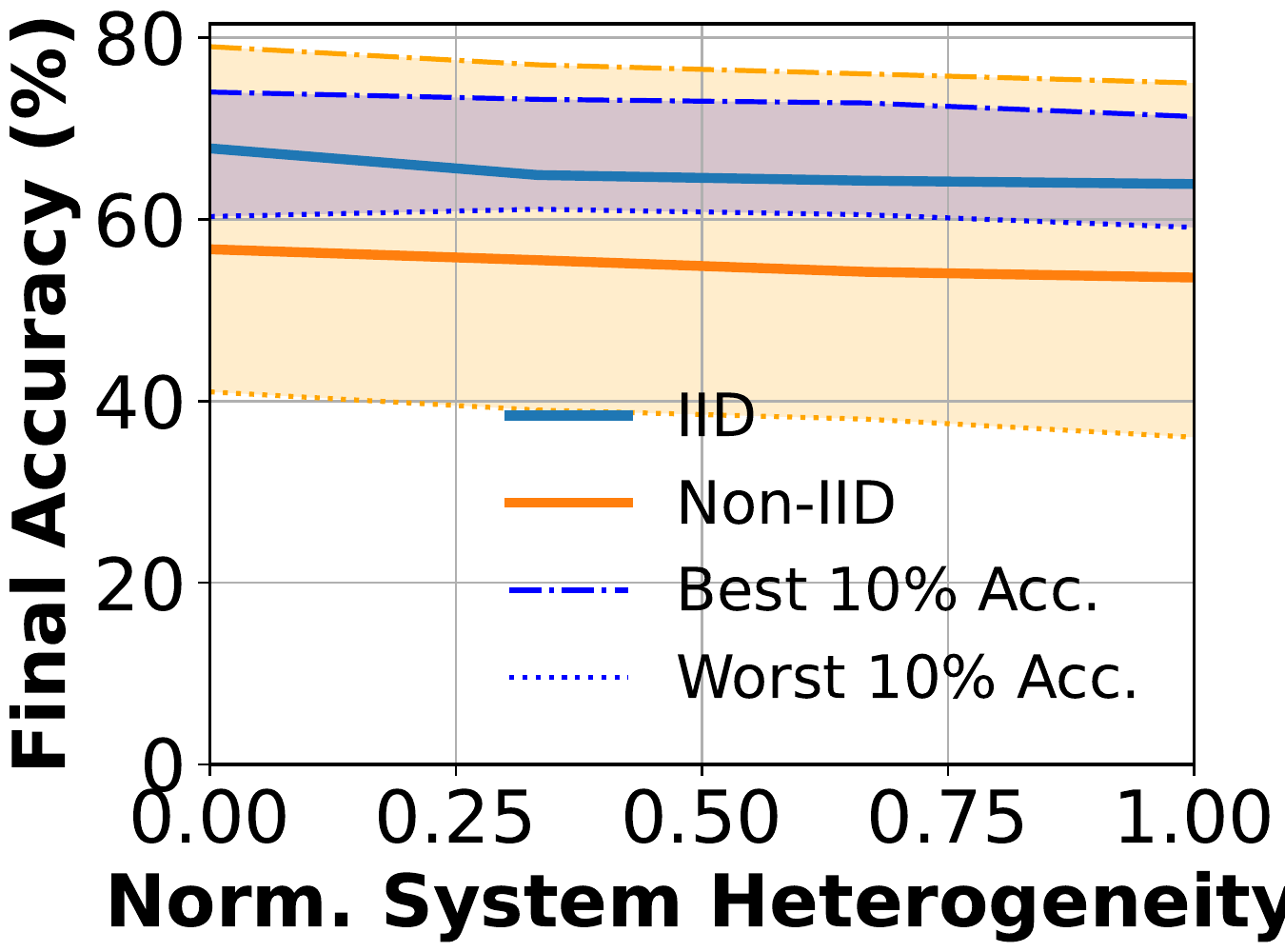} 
     \caption{System heterogeneity.  }
   \label{fig:sysheter}
     \end{subfigure}
    \caption{The impact of heterogeneity on final accuracy. }
\end{figure}
 
\paragraph{Impact of statistical heterogeneity.}
 Under large statistical heterogeneity across clients,  poor model accuracy, training time and fairness are often exacerbated, because the model is deployed on individual clients but is often trained over all the clients.  
Existing works that address statistical heterogeneity in FL assume bounded heterogeneity to simplify the problem complexity~\citep{convergence1, convergence2, yogi, prox}. 
However, we notice this does not hold in practical FL settings, which leads to great performance degradation under larger statistical heterogeneity.\footnote{In this experiment, we measure the statistical heterogeneity among a set of clients using the popular L2 distance on their data distributions~\citep{oort}.}
Indeed, our analysis of FedYoGi~\citep{yogi} (a state-of-the-art FL algorithm) on OpenImage~\citep{openimg} (an FL image dataset),
in Figure~\ref{fig:heter} shows that the model accuracy and its fairness across clients worsens with increasing statistical heterogeneity.
To achieve the same model performance under larger heterogeneity, more communication and/or computation costs are needed. 
This is true for personalization algorithms as well~\citep{personalization-cluster}. 
 
\paragraph{Impact of system heterogeneity.}
Heterogeneity of system-level characteristics raise challenges such as fault tolerance and straggler mitigation~\citep{prox,papaya}. 
Over-commitment~\citep{tff-paper}, which discards updates from slowest-responding participants, is commonly used to reduce the impact of stragglers, but it may lead to participation bias against slow devices.
Figure~\ref{fig:sysheter} shows the final accuracy of the OpenImage task under different degrees of system heterogeneity (variance of system speed).
For each experiment, we control the round duration and the number of successful participants to be the same; as a result, participation bias exacerbates with increasing system heterogeneity.
Since participation bias may enhance statistical heterogeneity in another form,
the final accuracy decreases with increasing system heterogeneity (albeit at a slower rate than statistical heterogeneity).

\begin{figure}[!t]
\centering
\begin{minipage}[t]{0.22\textwidth}
\centering
  \includegraphics[scale=0.3]{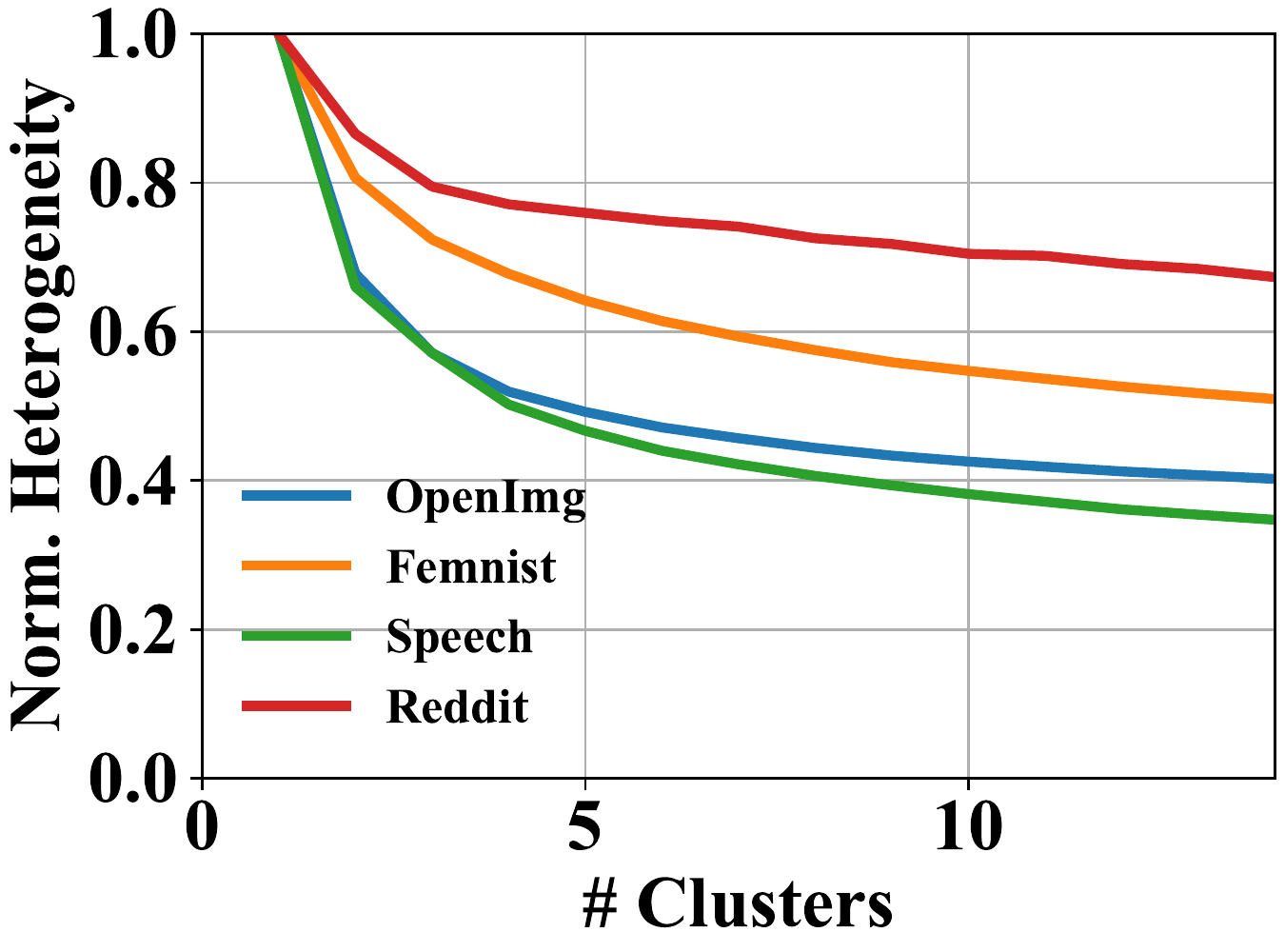} 
 \vspace{-1.5em}
    \caption{Intra-cluster heterogeneity in real datasets.}
  \label{fig:kmeans}
   \end{minipage}
      \hspace{2mm}
\begin{minipage}[t]{0.22\textwidth}
\centering
  \includegraphics[scale=0.3]{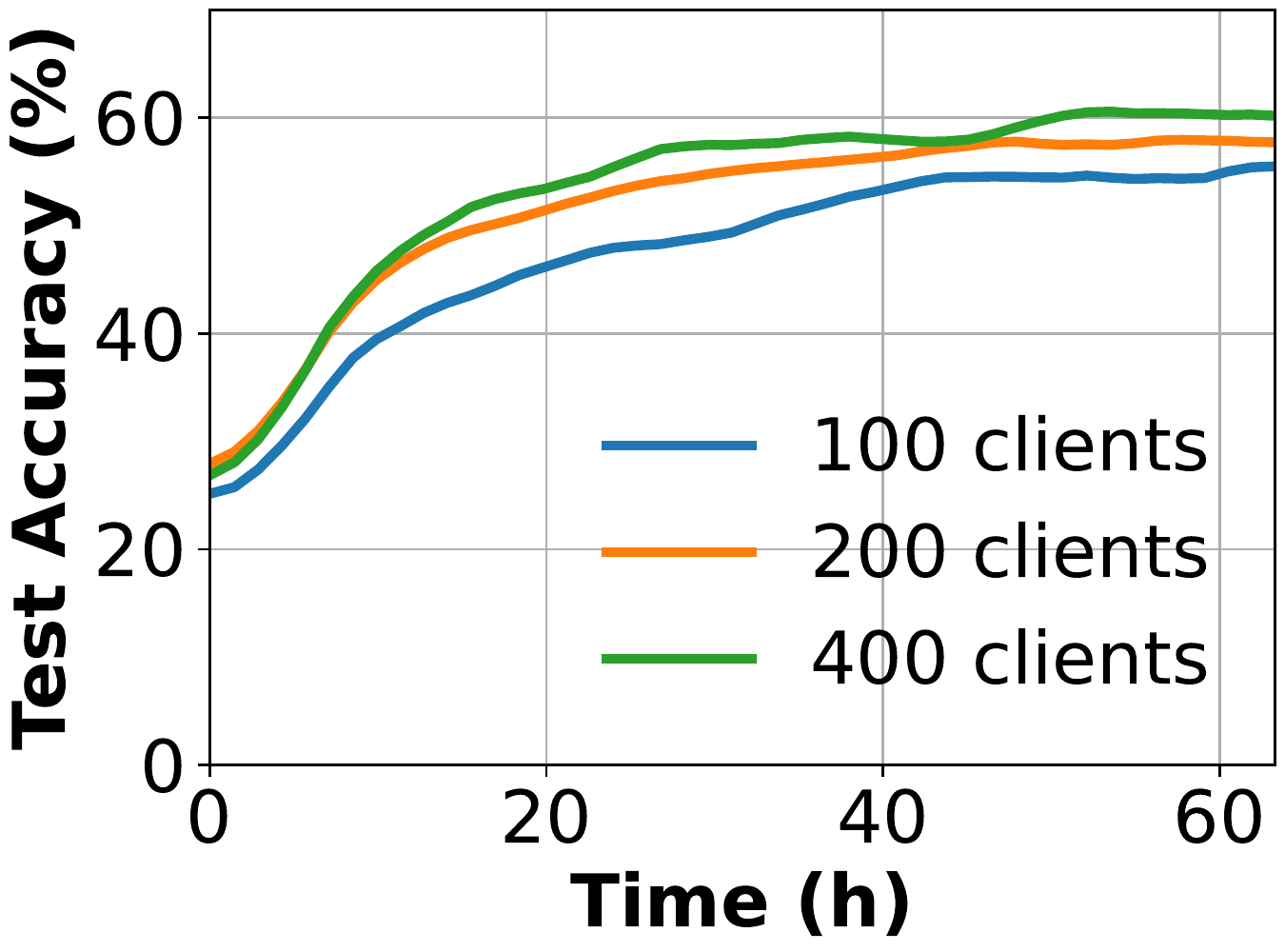} 
 \vspace{-1.5em}
    \caption{Diminishing return when adding more participants.}
  \label{fig:num_clt}
\end{minipage} 
\end{figure}
 
  \vspace{-.5em}

\subsection{Opportunities }
\label{sec:room-improv}
The opportunity for improving FL training performance, therefore, lies in decreasing heterogeneity especially the statistical heterogeneity based on the observation of the previous subsection.
By identifying statistically homogeneous groups and performing FL within each group, we may be able to boost model performance of most FL algorithms that are suffered by the heterogeneity.

Despite large statistical heterogeneity across the entire FL client population, there exist groups of statistically similar clients in most large populations.
Figure~\ref{fig:kmeans} shows that for four representative FL workloads~\citep{fedscale} in the real world.
We use K-means clustering (with increasing values of K) on clients' data distribution by their L2-distance metric. 
As the number of clusters increases from one (\ie, traditional FL with one global model) to larger values, we observe a small number of statistically similar groups emerge for most datasets.
 
 However, training K models to converge may need more training resources compared to training one model. 
 As shown in Figure~\ref{fig:num_clt}, increasing training resources has diminishing returns on the model convergence, which  presents the primary opportunity leveraged in this work: \emph{instead of letting all available clients contribute to a single global model, it may be more beneficial to partition them into several cohorts, each with smaller heterogeneity.}

%
 
\begin{table}[]
  \resizebox{\columnwidth}{!}{
  \begin{tabular}{c|ccccc}
  \hline
                                      & CFL      & FL+HC        & FlexCFL   & IFCA   & \name         \\ \hline
  Partial part.             & $\times$ & $\checkmark$ & $\checkmark$ & $\checkmark$ & $\checkmark$ \\
  Low avail.              & $\times$ & $\times$     & $\checkmark$ & $\checkmark$ & $\checkmark$ \\
  Res. constraint                 & $\times$ & $\times$     & $\times$     &  $\times$     & $\checkmark$ \\
  Training perf.    & $\times$ & $\times$     & $\times$     &$\times$     & $\checkmark$ \\
  \hline
  \end{tabular}}
   \caption{Comparing \name with existing Clustered FL.}
    \label{table:comparison}
\end{table}
  
\begin{figure*}[t]
  \vspace{-1cm}
  \begin{minipage}[t]{0.32\textwidth}
    \includegraphics[trim=0 70 0 100,  scale=0.95]{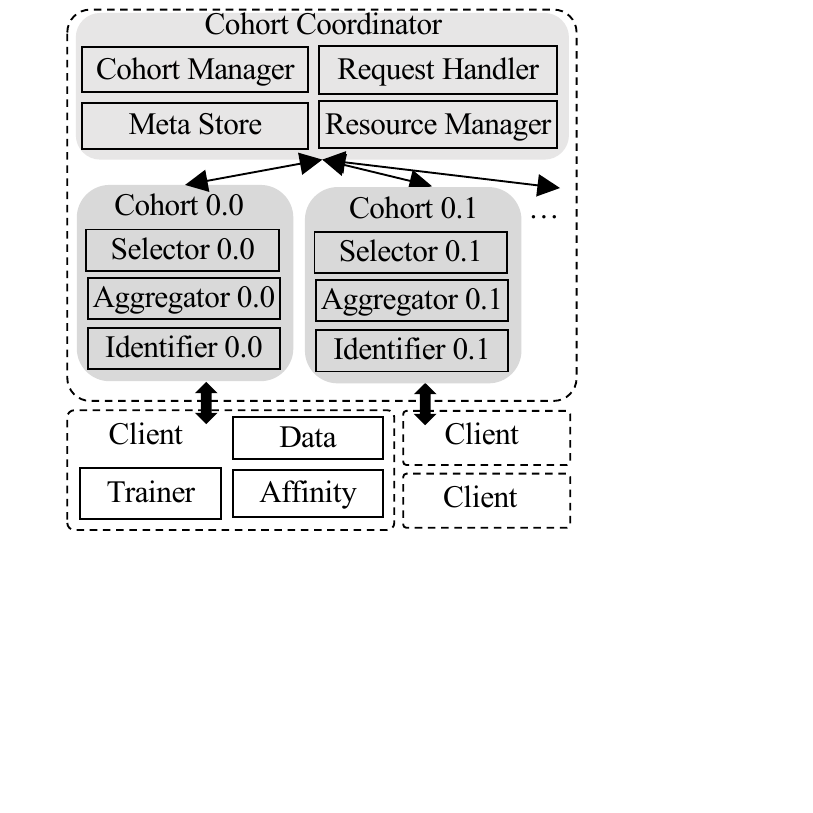} 
       \caption{\name architecture. \name server guides \name clients to train on their best-fit cohorts.}
     \label{fig:coco}
     \end{minipage}
        \hspace{1mm}
  \begin{minipage}[t]{0.66\textwidth}
  \centering
     \includegraphics[trim=0 70 0 0,clip, scale=1.2]{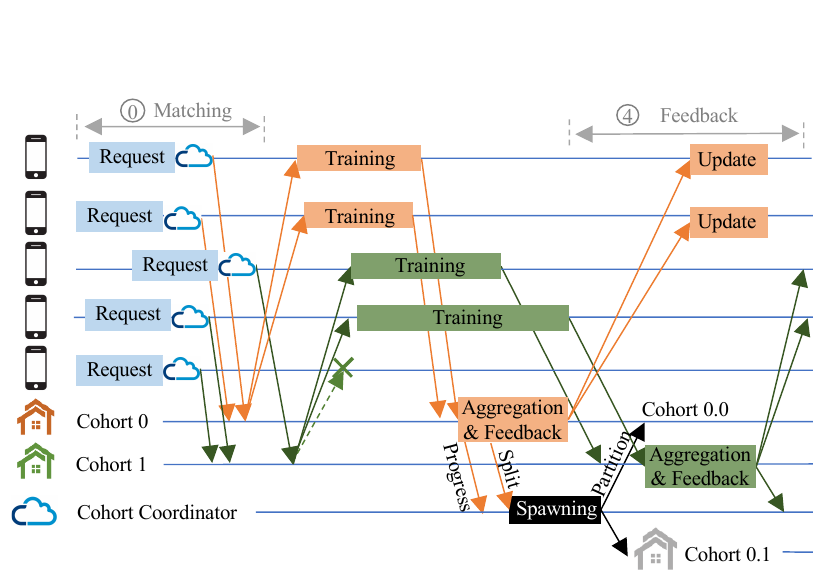}     
     \caption{
       {\name} lifecycle. 
       Clients check in with their affinity requests, and participate training within matched cohorts. 
       Cohorts are gradually identified based on participants response. }
     \label{fig:workflow}
  \end{minipage}
\end{figure*}

 \vspace{-.5em}

\subsection{Limitations of Existing Clustered FL }
\label{sec:challenge}
 
Recent efforts in the ML community have (theoretically) explored to create smaller groups of statistically similar clients. Yet, existing clustered FL algorithms often fall short across multiple dimensions in practical deployments, which motivates us to design systems support for efficient cohort identification and training. We empirically show the superior performance of \name over them too (\S\ref{sec:e2e}). 
 
\paragraph{Scalability.} 
FL in practice often involves millions of clients, and only a small fraction ($\sim$5\%~\citep{tff-paper, fedscale}) are available to participate in during a time window.
Such low availability and partial participation limit the available information for clustering algorithms. 
This, unfortunately, is ignored by CFL~\citep{cfl}, multi-center~\citep{multi-personal} and FL+HC~\citep{hc+fl}, making their deployment impractical as they require a complete pass over the entire population to identify clusters. 
Furthermore, clients usually have limited on-board resources, but IFCA~\citep{ifca}, FlexCFL~\citep{flexcfl}, ICFL~\citep{icfl}, k-FED~\citep{kfed} and FL+HC require extra computation for \emph{every} client to assign them to a cluster.
This imposes a significant computational and communication burden on already resource-constrained devices and diverts resources away from the primary task of model training.
For example, IFCA initiates multiple global models and broadcasts all models for each participant to choose from in each round; and FlexCFL and FL+HC require pre-training for every client to identify their clusters.

%

\paragraph{Efficiency.}
In addition to the challenge of identifying statistically similar groups at scale, how to leverage those similar groups to improve model performance introduces new trade-offs in deciding the right number of cohorts and time to partition.
Given a fixed amount of resources, generating more cohorts results in smaller heterogeneity; but it divides up the fixed training resource and unique training data per cohort, which hurts model convergence and generalizability. 
Moreover, partitioning clients too early can lead to model bias as the model is not generalized well by training on various clients, while partitioning too late can result in model variance over high heterogeneity. 
Unfortunately, most existing clustered FL algorithms are unaware of these tradeoffs, and rely on ad-hoc hyper-parameter tuning, which is prohibitively expensive as FL training can take many days and consume a large amount of resources.

In conclusion, as detailed in Table~\ref{table:comparison}, an effective client clustering solution in Federated Learning (FL) should take into account the following realistic constraints: 
\begin{enumerate}
  \item Partial participation: The algorithm should accommodate FL training that involves only a fraction of total participants in each round.
  \item Low availability: The algorithm should respect clients' sporadic availability, without necessitating participation from any clients at a specified time.
  \item Resource constraints: The algorithm should avoid demanding additional on-device computation for performing clustering.
  \item Training performance: The algorithm should optimize model performance—focusing on convergence and generalizability—within the constraints of a fixed training resource. This includes consideration of how clustering the FL population might positively impact performance despite reduced heterogeneity.
\end{enumerate}

%% file: pages/overview.tex
\section{\name OVERVIEW}%
\label{sec:overview}
 
\name progressively reduces the intra-group heterogeneity and improves the model performance through cohort identification and cohort-based training toward practical FL.
In this section, we introduce the cohort abstraction, provide an overview of how \name manages cohorts in a distributed fashion and fits into the FL life cycle.
 
\subsection{Cohort Abstraction}
\label{sec:cohort}
 
Instead of training only one global model, \name trains a model separately for each group of clients that shares  similar statistical data characteristics.
We refer to each of these groups, which can perform independent FL training over more homogeneous clients than the overall population, as a \emph{cohort} $C_m ( m\in [1,M])$ with two associated properties:
 
\begin{enumerate}
    \item A cohort should hold a specialized model that targets on it data distribution with smaller heterogeneity.    
    \item A cohort should have enough members $|C_m|$ to form a meaningful group and deliver the benefit of partition. 
\end{enumerate}
Traditional (\ie, cohort-agnostic) FL training has a single cohort with unbounded heterogeneity among the members.
 
\subsection{\name Architecture}
\label{sec:arch}

\name server consists of two primary components (Figure~\ref{fig:coco}):
 
\begin{enumerate}
  \item A logically centralized \textbf{cohort coordinator} performs three main functions. 
    First, it manages existing cohorts for fault tolerance.
    Second, it matches clients to their best-fit cohorts.
    Finally, it monitors the progress of cohort training and identification in order to decide cohort partition when it observes an opportunity for better model convergence .

  \item A set of \textbf{cohorts} each performs independent FL training.
    Each cohort contains traditional FL components such as aggregator and client selector.
    On top of traditional FL training activities, each cohort continuously identifies its internal composition,  reports its progress to the coordinator and waits for the partition instruction from the coordinator. 
\end{enumerate}

\paragraph{FL Lifecycle in \name}
 As shown in Figure~\ref{fig:workflow}, following the traditional FL stages in Section~\ref{sec:back}, \name adds a matching stage \circled{0} and a feedback stage \circled{4} before and after the traditional round.

\begin{enumerate}
  \item[\circled{0}] \textbf{Matching stage:} When checking in, clients using \name optionally include an affinity request (a hint about their cohort preference) to the cohort coordinator.
    If it took part in the training of one or more cohorts in the past, its preference is dependent on previous feedback. Otherwise, it has no preference.
    The cohort coordinator forwards the affinity request to the corresponding cohort based on its search algorithm and client's request.
 
  \item[\circled{1}]-\circled{3} \textbf{Traditional FL stages:} Each cohort starts a traditional FL training round independently after continuously receiving its client requests from the cohort coordinator. 
    These traditional stages include client selection, client training, server aggregation, and so on. 
    
  \item[\circled{4}] \textbf{Feedback stage:}
  After the traditional FL round finishes, each cohort updates the affinity feedback for its current participants based on the \name clustering algorithm (\S\ref{sec:algo}).
    Then, each participant receives an affinity feedback -- w.r.t. the cohort it trained with –- and updates the corresponding affinity record for submitting requests in a future round of FL training.
    During this stage, each cohort also reports its training and identification progress to the cohort coordinator.
 
\end{enumerate}
 
%
%
 
\textbf{Resource management:} 
\name jointly maximizes model convergence and resource efficiency in two ways.
First, its scalable cohort identification algorithm does not require extra on-device computation and uses the same amount of resources as traditional FL algorithms(\S\ref{sec:prob}- \S\ref{sec:cold-start}). 
Second, it carefully chooses the number of cohorts and time to partition to theoretically guarantee better model convergence and generalizability despite each cohort having less training resources than the previous global model (\S\ref{sec:system-tradeoff}).
  
%
%
%
%
%

  \textbf{Threat model and robustness.}
Like state-of-the-art production FL systems \citep{tff-paper,papaya, flint:mlsys23}, \name considers an \emph{honest-but-curious} centralized server for aggregation, which can infer any information without interfering with the FL training.
\name also assumes that most clients are honest (correct), and only a small fraction can act maliciously under the control of a bad actor \citep{FL-poisoning-attacks-response}. 
We elaborate on how \name can provide robustness under this threat model in Section~\ref{sec:robustness}.

%% file: pages/algorithm.tex
 \section{\name Clustering}%
\label{sec:algo}
 
In this section, we present the core clustering algorithm used in \name to identify cohorts (\S\ref{sec:prob}- \S\ref{sec:cold-start}). 
Then, we introduce the systems techniques to enable cohort-based training under realistic constraints (\S\ref{sec:system-tradeoff}).
 
\subsection{Problem Formulation and Overview}
\label{sec:prob} 
\name aims to accurately cluster clients by their statistical heterogeneity into appropriate cohorts under the following real-world FL constraints:  
\begin{enumerate}
\item \emph{Scalability: } The participants $\mathcal{P}^r$ in each round are only a small fraction of all clients ($N$), \ie, $|\mathcal{P}^r| \ll N $. How to identify cohorts and cluster clients at scale under such low client availability?
\item \emph{Resource Efficiency: } How to conduct the clustering process without incurring overhead on devices, such as extra model training and client participation that do not contribute to model training? 
\item \emph{Information Deficiency: }  The information available to today's FL central server is limited to such as gradients and training loss. How to cluster clients without requesting additional information from clients?
\end{enumerate}
\paragraph{Problem Formulation: }
The \textit{input} to the server is a list of participants along with their gradients collected over training rounds based on these two constraints. 
Intuitively, the gradient of client relies on its local dataset $x_i$ and the received model weights (unique for the round $r$ and cohort $m$), and this gradient is multi-dimensional,  embedding more information than its counterparts (\eg, training loss). 
As such, we can formulate the \textit{input} of the clustering algorithm in each round $r$ as $\{\{g_m^r(x_i) \}_{ i \in \mathcal{P}_m^r} \}_{m\in [1, M_r]}$, where $g_m^r(x_i)$ is the gradient of participant $i$ , $\mathcal{P}_m^r$ is the participants list, and $M_r$ is the number of cohorts.
 
The \textit{output} is the cohort membership $\{S_i \in [1, M]\}$ for each client $i\in[1,N]$. Following the objective of traditional clustering algorithms~\citep{kmeans-def}, \name also aims to minimize the average intra-cohort heterogeneity ($J$) defined as:

   \begin{equation}
  J = \sum_{m=1}^M \frac{1}{2|\{x | S_x=m\}|} \sum_{S_i,S_j = m} || {x_i} - {x_j} ||^2.
   \label{eq:obj}
\end{equation}

Intuitively, we can model it as a clustering problem $\{ x_1, ... , x_N \} \rightarrow \{ S_1,...,S_N  \}$, whereas doing so encounters new challenges.
\begin{enumerate}
\item  How to derive client data similarity without direct access to data and without iterating all but part of the clustering objects every round.
\item How to assign new incoming clients to the best-fit cohort without prior information after \name generates more than one cohorts.
\end{enumerate}
 
  Following this problem definition and challenge, Algorithm~\ref{algo:all} illustrates the overview of \name clustering mechanism, 
which consists of an online cluster algorithm to cluster clients at scale (\S\ref{sec:clustering}) and the cohort selection for individual FL clients (\S\ref{sec:cold-start}). 
Note that, \name's clustering algorithm can operate in the background, imposing no additional overhead on the training process.

\SetNlSty{large}{}{:}
\SetKwProg{Fn}{Function}{:}{}

\begin{algorithm}[!t]
  \DontPrintSemicolon

  \SetAlgoLined
  
     \textbf{  Input: }{Participants list $\mathcal{P}$, Exploration factor $\epsilon$} 
     
      \textbf{  Output: }{Client-cohort membership list $S_{\mathcal{D}}$}
      
   $M \leftarrow 1$;  
 \algcomment{Initialize the number of cohorts.} 
 
   $S_{\mathcal{D}} \leftarrow 0 $;  
 \algcomment{Initialize client-cohort membership.} 
 
  $R_{\mathcal{D}, M } \leftarrow 0 $;   
\algcomment{Initialize client-cohort reward.}  

           $L_{\mathcal{D}, M } \leftarrow   N/A. $  
\algcomment{Initialize client-cohort cluster id.}


         \For{each round $ r = 1, 2, . . .  $}         {
           
             $\mathcal{P}_{m}^r = 	\{ i | S_i = m, i \in \mathcal{P}^r$\} \label{code:reward}      
             
         \For{each cohort $ m = 1, . . . , M $  in  parallel }  {
		 $R_{\mathcal{P}_m^r}$ =  \texttt{\name-Clustering} ($\mathcal{P}^r_m  $  )}

               $S_{\mathcal{P}_m^r}= \texttt{CohortSelection}  (R_{\mathcal{P}_m^r}, \epsilon, r )$                   \label{code:select}           
     }
  
     \Return $S_{\mathcal{D}}$
\vspace{2mm}

\SetKwFunction{ClientClustering}{\textrm{ClientClustering}}
\Fn{\ClientClustering{Participants list $\mathcal{P}^r_m $}} {  \label{code:func}

   \tcc{Identify clusters on the fly.  (\ref{sec:system-tradeoff}) }
 	\If{$r==1$}{ \label{code:kmeans}
             $L_{ \mathcal{P}_{m}^r, m } = \texttt{Kmeans}( g_{m}^r(x_{\mathcal{P}_{m}^r}), K).  $ \label{code:kmeans-label}}
   	\Else {\label{code:assign}
	
	   $\mathcal{P}_k = \{i | L_{i,m} = k, i\in \mathcal{P}_{m}^r\}, \forall  k \in  [0,K)$ \label{code:cluster-cen1} 
        
         $ C_k = \overline{\{ g_m^r(x_{\mathcal{P}_j}) }, \forall  k \in  [0,K)$   \label{code:cluster-cen2}
    	 
	  $ L_{\mathcal{P}_{m}^r,m}  = arg\min_k ||  g_m^r(x_{\mathcal{P}_m^r}) - C_k ||_2	$  \label{code:label}
        } 
         
          \tcc{Decide partitioning to start separate training.  (\ref{sec:clustering})}
  	\If{ \texttt{PartitionCriteria}(m) } {\label{code:decision}
	
	  $ M=M+K-1$
	  
	  $R_{\mathcal{D}, m+k} = R_{\mathcal{D}, m}+0.1*\mathds{1}(L_{\mathcal{D},m}==k), \forall  k\in [ 0,K)$ 
  	} 
	
     \tcc{Update rewards for cohort selection. (\ref{sec:system-tradeoff}) }
  \If{$M > 1$}{    \label{code:assign}
           $  R_{ \mathcal{P}_{m}^r ,m } =\texttt{ExploitReward}(  R_{ \mathcal{P}_{m}^r ,m }, x_{\mathcal{P}^r_m} ) $ \label{code:exploit}
     
         $ R_{ \mathcal{P}_{m}^r ,m' } $=\texttt{ExploreReward}($ R_{ \mathcal{P}_{m}^r } ,m'$),  $\forall m \neq m'$      \label{code:explore}
       
       \label{code:efficient}
  }
           
     \Return $R_{\mathcal{P}_m^r}$
}
 
\vspace{2mm}

\SetKwFunction{CohortSelection}{\textrm{CohortSelection}}
\Fn{\CohortSelection{ (Reward list $R_{\mathcal{P}_m^r}, \epsilon, r $)} }{  \label{code:func}

\For{client i in $\mathcal{P}_m^r$}{
\If{random(0,1) > $\epsilon^r$}{
$S_i $=random(0, M)
}
\Else{
$S_i = \arg \max R_i $

}

}
     \Return $S_{\mathcal{P}_m^r}$
     
     }

    \caption{\name Clustering Algorithm  }
    \label{algo:all}
    \label{alg:overview}
\end{algorithm}

%
%
%
 
\subsection{Online Clustering}  
\label{sec:clustering} 
\name resorts to the similarity of clients' gradients to capture their statistical similarity. Our design is inspired by the recent advances in ML theory~\citep{cfl,proof}, 
which show that the data heterogeneity can attribute to the gradient divergence~\citep{prox} and a smaller heterogeneity would have smaller gradient divergence for the \emph{same} initial model weight. 
Here, we measure such gradient divergence using the widely-used cosine similarity~\citep{cosine} among the input batch of gradients $g_m^r(x_i), i \in \mathcal{P}_m^r$ to investigate client similarity.\footnote{Cosine similarity measures the similarity between two vectors of an inner product space~\citep{cosine}.} Compared to other counterparts such as L-2 distance which does not take into account the direction of the gradients, cosine similarity better quantifies how similarly their needed model changes are directed.
 
However, the sporadic participation of clients in each training round limits the data available for clustering algorithms to a subset of the entire client population at any given time. 
Traditional clustering algorithms, such as K-means and KNN, require a complete pass of the population, rendering them inapplicable here. 
Mini-batch clustering algorithms~\citep{mini-batch}, on the other hand, operate on small batches of the population each round, maintain a running centroid for cluster assignments.
Nonetheless, this strategy cannot directly be applied in our case because we only know the gradients ${g_m^r(x_{\mathcal{P}m^r}) }$ and not the raw data $x_{\mathcal{P}}$. Further, since the gradient $g_m^r(\cdot)$ depends on the initial model of round $r$ and client data - both unknown and different across rounds and cohorts.
These complexities preclude us from maintaining absolute cluster centroids over successive rounds in a straightforward manner, making naive mini-batch clustering infeasible.

Algorithm~\ref{algo:all} outlines how \name starts with one cohort for the entire FL population, and then adaptively identifies cohorts based on gradients of mini-batch clients.
After using K-means to initialize the cluster prototype (Line~\ref{code:kmeans-label}), in each round, \name collects the training feedback from the clients and assigns clients to their closest clusters (Line~\ref{code:cluster-cen1}).
Meanwhile, \name incrementally refines cluster centers based on the gradients of newly assigned clients in each round (Line~\ref{code:cluster-cen2}).
 With repeated cluster updating and clients assignment, \name can effectively identify the clusters at scale (Line~\ref{code:label}).
 Each new cohort starts with the parent cohort model weights with the same architecture, performs conventional FL steps separately, and converges to different model weights. 
Once discernible clusters emerge and certain partition criteria are fulfilled (\eg, enough participants left for model convergence after partition), \name decides to spawn cohorts based on these pre-identified clusters (Line~\ref{code:decision}) and train cohort models separately within their corresponding client groups. 
At runtime, \name adaptively decides the right time and the right number of cohorts to partition to find the sweet spot of model performance and the resource consumption of training multiple cohorts (\S~\ref{sec:system-tradeoff}).

\subsection{Cohort Selection}  
\label{sec:cold-start}

Although clustering captures the membership of already-identified clients, 
doing so for a new client is unkown a priori, since we neither have access to client data nor have absolute cohort centers that can inform a new client to choose the closest cohort. 
This challenge is further amplified by the large training population, 
wherein more FL clients participate in model training for the first time than not. 

To address this, \name adopts an \emph{exploration-exploitation} strategy to efficiently identify the cohort membership for new participants (Line~\ref{code:select}).
This allows us to first randomly assign a new client to a cohort. 
After getting the feedback on how well the client fits in that cohort, \name attempts to identify a more suitable cohort for it the next time it participates again.

\name uses reward-based decaying $\epsilon$-greedy selection~\citep{mab} to help the client find the best-fit cohort (Line~\ref{code:select}). 
With an aim to maximize the expected reward for each client, there is a $1- \epsilon$ probability of selecting a cohort with a maximum reward and a $\epsilon$ probability of selecting cohorts randomly, where $ \epsilon \in[0,1] $ is the exploration factor that decays over time to account for the latest information. 
Intuitively, smaller gradient divergence compared to the members within the explored cohort means a better fit and gives a higher reward. 
Hence, \name calculates the relative divergence between the client gradients and the explored cohort center.
 This is done by first estimating the cohort center via averaging the client gradients within the cohort $\mathcal{P}_{m,Known}^r$, to be $ D = || g_m^r(x_{\mathcal{P}^r_m})  - \overline{  g_m^r(x_{\mathcal{P}^r_{m,  Known}})} ||_2$, where $ \overline{  g_m^r(x_{\mathcal{P}^r_{m,  Known}})} $ represents the estimated cluster centers for cohort $m$. 
Next, we take the popular approach to identify outlier clients~\citep{outlier}.
Specifically, we consider clients as outliers if their distance to the cohort center exceeds the threshold, which is calculated as the sum of the mean and the standard deviation of $D$, denoted as $avg(D) + std(D)$.
If the client gradient distance to the cohort center is larger than this threshold, this client is not considered as the cohort member.
As such, the instant reward becomes $ \Delta R  =  1 - \frac{1}{avg(D) + std(D)} D$, where the client with a negative $\Delta R$ would be considered as an outlier of the cohort.  
Then, \name updates the reward between each client and its explored cohort with a decay factor $\gamma$ as  $  R_{ \mathcal{P}_{m}^r ,m } =  \gamma *\Delta R + (1-\gamma)*  R_{\mathcal{P}_{m}^r, m}  $, $\gamma$=0.2 by default in popular exploration-exploitation designs.

\paragraph{Efficient cohort exploration.}
During exploration, there may exist multiple cohorts for a client to try out with. 
To improve the searching efficiency and save device training resources, during both training and deployment, 
\name enables a new client to perform a binary search to find the most appropriate cohort by predicting the rewards for other unexplored cohorts $m'$ through function \texttt{ExploreReward()} (Line~\ref{code:explore}):
$
  R_{ \mathcal{P}_{m}^r ,m' } \mathrel{{+}{=}} \frac{  R_{ \mathcal{P}_{m}^r } }{d(m,m')+1} , \forall m \neq m'.
 $

The intuition behind the cohort search is that the client may perform similar to or receive similar rewards from the cohorts that are closer/similar to the previously explored ones, and vice versa.
To find out the cohort similarity, we first define the distance ($d$) between two cohorts to be the distance to their lowest common ancestral cohorts based on the hierarchical cluster relationship among cohorts.
Given an explored cohort $m$ and the reward $\Delta R_m$ for a participant, \name calculates the distance $d$ and updates the rewards for unexplored cohorts to be inversely proportional to their distance.
 For example, if a client receives a negative reward for the chosen cohort, then he is more likely to explore another furthest cohort with higher reward given by \texttt{ExploreReward()} next time. 
  

\begin{figure}[t!]
\vspace{-0.2cm}
   \centering
   \includegraphics[trim=0 90 0 0,clip,scale=0.9]{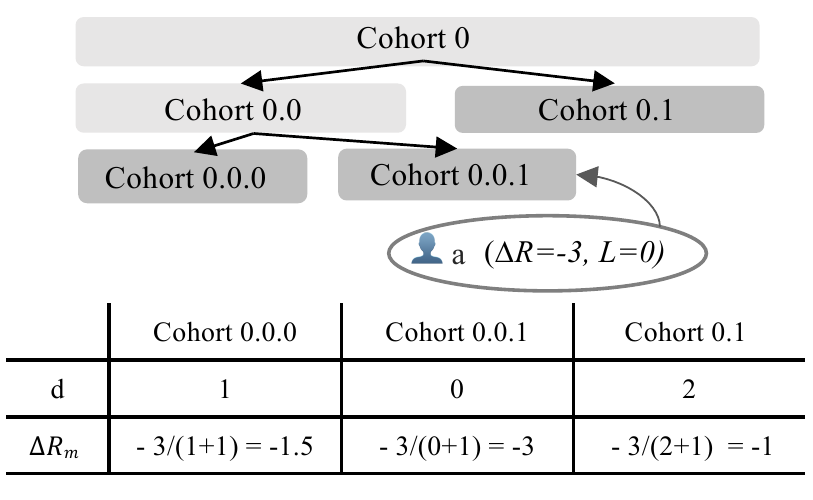} 
     \caption{Reward update based on the hierarchical structure among all cohorts.}
   \label{fig:toy}
\end{figure}

Taking Figure~\ref{fig:toy} as an example, a new client $a$ explores  $Cohort_{0.0.1}$ and receives the corresponding feedback rewards -3.
Then, with the intuition that the client may have similar performance on a closer cohort, \name calculates the distance between $Cohort_{0.0.1}$  and other cohorts.
As shown in Table~\ref{fig:toy}, \name updates the rewards to be inversely proportional to the cohort distance $d_{m'}$: 
$\Delta R_{m'}  =  \frac{\Delta R}{d_{m'}+1} $
Since $Cohort_{0.0.1}$ and $Cohort_{0.1}$ have a larger distance between them, $Cohort_{0.1}$ ends up with a relatively higher reward and has higher probability to be explored by client $a$ in the future.

\subsection{Cohort-Based Training}
\label{sec:system-tradeoff}
   
  While clustering reduces heterogeneity within a cohort, 
  generating a larger number of cohorts may dilute available resources for each individual cohort when operating under fixed training resources.
   Consequently, this leads to a new trade-off between resource efficiency and model convergence. 
As shown in Figure~\ref{fig:kmeans}, generating more cohorts has diminishing returns in terms of heterogeneity. 
In a setting where total training resources are fixed, allocating resources to a larger number of cohorts implies fewer resources for each, which may negatively affect model convergence. 
Conversely, having too few cohorts is insufficient for adequately addressing intra-cohort heterogeneity. 
Therefore, \name faces the challenge of optimally determining both the number of cohorts and the timing for their creation to balance resource efficiency and model performance effectively.

Intuitively, the decision to generate new cohorts should be based on the extent of client heterogeneity and the available training resource budget post-partition. When client heterogeneity is significant and the resource budget is sufficient, the creation of additional cohorts is warranted to further reduce client heterogeneity. On the other hand, when these conditions are not met, the creation of new cohorts should be deferred.

We next provide analytical insights to ground our strategy.
Prior works in ML theory~\citep{scaffold, convergence1, convergence2, yogi} have shown that 
the convergence rate of FL training is largely dominated by heterogeneity. 
We start by analyzing the relationship between heterogeneity and training resources in theory. Inspired by the convergence analysis of FedAvg~\citep{scaffold}, we establish the following Lemma. More detailed proof are available in Appendix~\ref{sec:proof}.

\begin{lemma}
 If the population and training resources are partitioned into up to $K$ cohorts, to theoretically achieve better model convergence, intra-cohort heterogeneity should be reduced by $\sqrt{K}$ times 
when the training resource $|\mathcal{P}|$ is larger than $\alpha \sqrt{ \frac{  |\mathcal{P}_0| }{J_0^2}}$.
  $\alpha$ is a constant setting that elaborates the relationship between model convergence and training resources.
\label{lm:res}
\end{lemma}

From Lemma~\ref{lm:res}, we notice that the number of generated cohorts rely on the expected reduced heterogeneity and a lower bound of training resources.
 As such, \name actively monitors the gradient divergence within each cohort at runtime to estimate the potential heterogeneity reduction. 
 
 When a sufficient decrease (\eg, $\frac{1}{\sqrt{K}}$) in intra-cohort heterogeneity and ample post-partition training resources are detected, \name autonomously partitions the population into a maximum of $K$ cohorts, allotting equal training resources to each. 
 This strategy theoretically enhances model convergence through cohort-based training in \name. 
As for some FL datasets with larger heterogeneity, FL developers can further improve model convergence by dynamically raising the resource budget to allow generating more cohorts.

In addition to deciding the right number of cohorts, the time to cohort partition is also critical to model convergence.
As cohort partitioning may reduce the unique training data for each cohort model, the trade-off between model bias and variance can be affected by the time of partition.
On the one hand, hard partitioning of the entire population at the beginning could reduce heterogeneity, but it could also reduce the amount of unique training data for each cohort model, leading to poor model generalizability.
On the other hand, late partitioning exposes the model to diverse training data but leads to worse model variance due to high heterogeneity.
These also guide the \emph{reuse} of identified cohorts to facilitate other FL tasks. 

From the sensitive analysis of cohort partition time (\S\ref{sec:partition-time}), we found the model convergence is not sensitive to exact partition time as long as cohorts are not partitioned at the beginning or the end of the training. 
We report more results about the effect of partition time on model convergence in Section~\ref{sec:partition-time}.

 Finally, the start time of gradient-based clustering can impact the efficiency of the process. 
In the early stages of training, gradients are often large and may not adequately capture the distributional features of the data. 
However, as the model approaches convergence, the gradients become more informative indicators of data similarities. 
Thus, it is crucial for \name to judiciously select the optimal starting point for clustering so as not to delay the cohort identification. 
Detailed results discussing the effect of the clustering start time on model convergence can be found in Section~\ref{sec:partition-time}.

%% file: pages/sys.tex
 \section{\name System Design}%
\label{sec:system}

In this section, we discuss how to design a practical and robust system on top of the clustering algorithm under real-world challenges.

\subsection{Distributed \name}
\label{sec:dist} 

As the scale of training grows, the server faces more server challenges for tremendous storage, fault tolerance, and client privacy in order to maintain the cohort and client information.
Thus, \name designs a solution to use a soft-state server that offloads cohort-related information to individual clients to mitigate these challenges.
In this subsection, we describe how to implement the proposed clustering algorithm in a distributed fashion, while achieving the same objective.

Firstly, we introduce \emph{affinity message}, which is a lightweight message containing all necessary 
state information needed to identify cohorts in a distributed fashion.
Affinity message consists of  two pieces of information between a client and a cohort  to enable efficient state transmission: (\emph{Reward} $R\in \mathbb{R}$, \emph{Cluster index} $L\in [0, K)$).
The \emph{reward} implies how well the client fits for this cohort.
The \emph{cluster index} expresses the client's cluster membership within this cohort and is used to indicate its future cohort index. 

\begin{figure}[t!]
  \centering
  \includegraphics[trim=0 160 0 0,clip,scale=1]{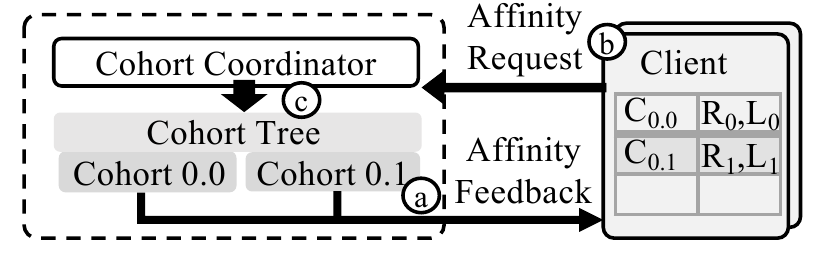} 
    \caption{Scalable Design Overview
  \textbf{a.} Cohort affinity feedback.  \textbf{b.} Client affinity request.  \textbf{c.}  Cohort coordinator request match.
    }
  \label{fig:stateless}
\end{figure}

Through exchanging affinity messages between different components, \name encourages similar clients to collaborate more in a distributed fashion.
As shown in Figure~\ref{fig:stateless}, we next describe
\circled{a} how a cohort informs its relationship with its participants, 
\circled{b} how clients request for their preferred cohort based on the affinity feedback
and \circled{c} how the cohort coordinator matches different requests.

\paragraph{Affinity Feedback}

At the end of each round, every cohort computes the affinity feedback to inform participants about their relationship with the cohort.
These affinity feedback correspond to the clustering results returned by Algorithm~\ref{algo:all} Line~\ref{code:reward} (reward $R$) and Line~\ref{code:label} (cluster index $L$).
These clustering results would be sent back to the participants respectively in the format of affinity messages,
which informs the participant about whether the cohort is a good fit and which sub-cohort to select after partitioning.

\paragraph{Client Reaction }

After receiving the affinity feedback from the cohort, the client would update its affinity records itself based on the equation in Algorithm~\ref{algo:all} Line~\ref{code:exploit}-~\ref{code:explore}
 and copy the cluster index directly.
Following the same decaying $\epsilon-$greedy selection method (\S\ref{sec:cold-start}), clients select the cohort to train by themselves.
Then, clients ready to participate would submit the corresponding affinity request to the cohort coordinator.

\paragraph{Request Match} 

After receiving the affinity request,  the cohort coordinator matches each client to the cohort it requests.
Note that only the leaf cohort in the cohort tree would be returned as it conducts actual FL training inside.
The requested cohort may not be the leaf cohort because some clients may not be aware of the cohort partitioning, which is not yet transparent to all clients. 
In this case, the cohort coordinator should assist clients to select their best-fit cohort through finding the closest leaf cohort indicated by the requested cohort and cluster index in the affinity message.

After finding a proper cohort, cohort coordinator would forward this affinity request to the corresponding cohort to initiate traditional FL rounds. 
Moreover, these forwarded affinity requests provide each cohort with all necessary input to conduct the clustering algorithms.
Thus, after receiving the gradients from its participants, each cohort is able to run the Algorithm~\ref{algo:all} independently to compute the aforementioned affinity feedback. 

\subsection{Resilient \name}
\label{sec:resilience}

\paragraph{Fault Tolerance}

\name enables fast recovery to minimize the impact on the model training.
Upon a cohort process failure in the server, the cohort coordinator spawns a new cohort process.
The new cohort loads the model from the latest checkpoint and restarts the incomplete round.

If the cohort coordinator fails, cohort processes would continue their current independent FL training and wait until a new cohort coordinator to be re-spawned.
Clients checking in within that recovery period would be ignored.

Finally, \name is resilient to client failures just like traditional FL by design. 
Most client failure handlers, which are orthogonal to \name, can be applied directly.
In addition, a failed client, who may lose its own affinity records, would restart exploring again.
We empirically show that \name can tolerate a certain amount of such client failures while continuing to benefit the FL training (\S\ref{sec:dp}).

\paragraph{Robustness}
\label{sec:robustness}

Based on \name's threat model, \name can naturally cooperate with some existing privacy-preserving approaches~\citep{ldp,dprnn,recldp} to address potential threats from both the server and clients.
To handle the honest-but-curious server, \name can be used with local differential privacy (LDP), which is used to provide user-level privacy guarantees.
Since differential privacy is immune to post-processing~\citep{dpbook} and \name's clustering is post-processing, \name incurs no additional privacy loss.

To handle a small fraction of unreliable clients~\citep{malicious1,malicious2}, \name can be used with existing adversary-resilient solutions~\citep{resilient4,resilient3}. 
For \name-specialized adversaries, such as fake affinity requests,
\name detects anomalies by comparing its position in the cluster with its claimed affinity  (Algorithm~\ref{algo:all} Line~\ref{code:reward}). 
If a significant discrepancy is detected, \name will detect and blacklist it.
In Section~\ref{sec:dp}, we empirically evaluate \name's robustness under these scenarios.

%% file: pages/impl.tex
\section{IMPLEMENTATION}
\label{sec:impl}

We have implemented \name as an independent Python library ($1,664$ lines) to serve existing FL frameworks (e.g., TFF~\citep{tff} and PySyft~\citep{pysyft}), and integrated it with FedScale~\citep{fedscale} for evaluations.
\name abstracts away the cohort identification and partition so that FL developers can easily try out their FL algorithms or datasets on top of \name without any modifications.

\name's implementation consists of the three components described in Section~\ref{sec:overview}: 
The cohort coordinator manages and spawns cohort processes, which initiate FL training tasks.
Clients continuously submit their training requests based on their availability and affinity records.
Then, the cohort coordinator takes client training requests as input and forwards the requests to corresponding cohorts.
Each cohort process conducts conventional FL training with the assigned available clients independently.
At the end of each individual round, the \name clustering algorithm runs within every cohort and reports clustering results to each participant over the network.
All training metadata and model weights are checkpointed periodically for fault tolerance.
Meanwhile, the cohort coordinator continuously monitors the progress of cohorts for resource management and failure recovery.

%% file: pages/eval.tex
\section{EVALUATION}%

\label{sec:eval}
\begin{table}[t!]
  \small
  \begin{center}
  \begin{tabular}{crr}
  \toprule 
  Dataset	& \#Clients	& \#Samples \\
  \midrule 
      Google Speech~\citep{google-speech}	& 2,618	&	105K	\\
      FEMNIST~\citep{FEMNIST}	&	3,400	&	640K			\\
      OpenImage-Easy 		&	10,133	&	1M		\\	
      OpenImage~\citep{openimg}		&	13,771	&	1.3M		\\
      Amazon Review~\citep{amazon}	&  42,031	&	2M		\\
      Reddit~\citep{reddit}	&  63,058	&	5M			\\
  \bottomrule 
  \end{tabular}
  \end{center}
 \caption{Statistics of the six datasets in evaluation.}
                \vspace{-1.5em}

  \label{table:data-stats} 
\end{table}
\definecolor{Gray}{gray}{0.9}
\newcolumntype{g}{>{\columncolor{Gray}}c}
\renewcommand{\arraystretch}{1.3}
\begin{table*}[ht]
\centering
\begin{tabular}{ccccgg}
\hline
Task                                   & Dataset                        & Model                                                                                            & Target Acc. & \cellcolor{white} \name Speedup    & \cellcolor{white} \name Acc. Impr.   \\ \hline
                                       & FEMNIST                        & ResNet-18                                                   & 82.2\%            & $1.2\times  $                     & 7.3\%                                   \\ \cline{2-6} 
                                       &                                & MobileNet     & 56.5\%            & $   1.3\times  $                     & 4.8\%                                   \\ \cline{3-6} 
                                       & \multirow{-2}{*}{OpenImg}      & ShuffleNet      & 58.2\%            & $   2.2\times  $                     & 5.0\%                                   \\ \cline{2-6} 
                                       &                                & MobileNet                                        & 65.4\%               & $ 1.4 \times  $                       & 3.4\%                                      \\ \cline{3-6} 
\multirow{-5}{*}{Image Classification} & \multirow{-2}{*}{OpenImg-Easy} & ShuffleNet                                       & 64.8\%               & $1.2\times  $                       & 4.4\%                                      \\ \hline
                                       & Amazon Review                  & Logistic Regression                                                                              & 65.3\%               & $ 1.2 \times   $                      & 8.2\%                                      \\ \cline{2-6} 
\multirow{-2}{*}{Language Modeling}    & Reddit                         & Albert                                             & 7 perplexity      & $   1\times  $                     & 0 perplexity                                    \\ \hline
Speech Recognition                     & Google Speech                  & ResNet-34                                                                                        & 78.5\%            & $   1.5\times  $                     & 5.7\%                  \\ \hline
                     
\end{tabular}
  \caption{Summary of improvements on time to accuracy. 
We target the highest accuracy attainable by YoGi.    
    } 
  \label{table:e2e-perf}
\end{table*}

\renewcommand{\arraystretch}{1}
 
We evaluate \name's effectiveness for six different ML tasks as well as different choices of FL  algorithms.
Our evaluation shows the following key highlights:
\begin{enumerate} 
\item \name  speeds up model convergence on different FL datasets up to 2.2$\times$, while improving final test accuracy by 3.4\%-8.2\%.  
 \name  cooperates with existing FL efforts (\eg, personalization) and boosts final test accuracy by 2.1\%--6.7\%.
 \name can mitigate model bias across devices by 4.8\% and 53.8\% and improve resource efficiency (\S\ref{sec:e2e}).
\item \name outperforms existing clustered FL solutions up to 4.8$\times$ in time and 5.2$\times$ in resources (\S\ref{sec:exp-cfl}).   
\item \name performs well across a broad range of its parameter settings (\S\ref{sec:partition-time}).   
\end{enumerate}
   
\begin{figure*}[t!]
  \centering
 \begin{subfigure}{1.62in}
    \centering
    \includegraphics[scale=0.3]{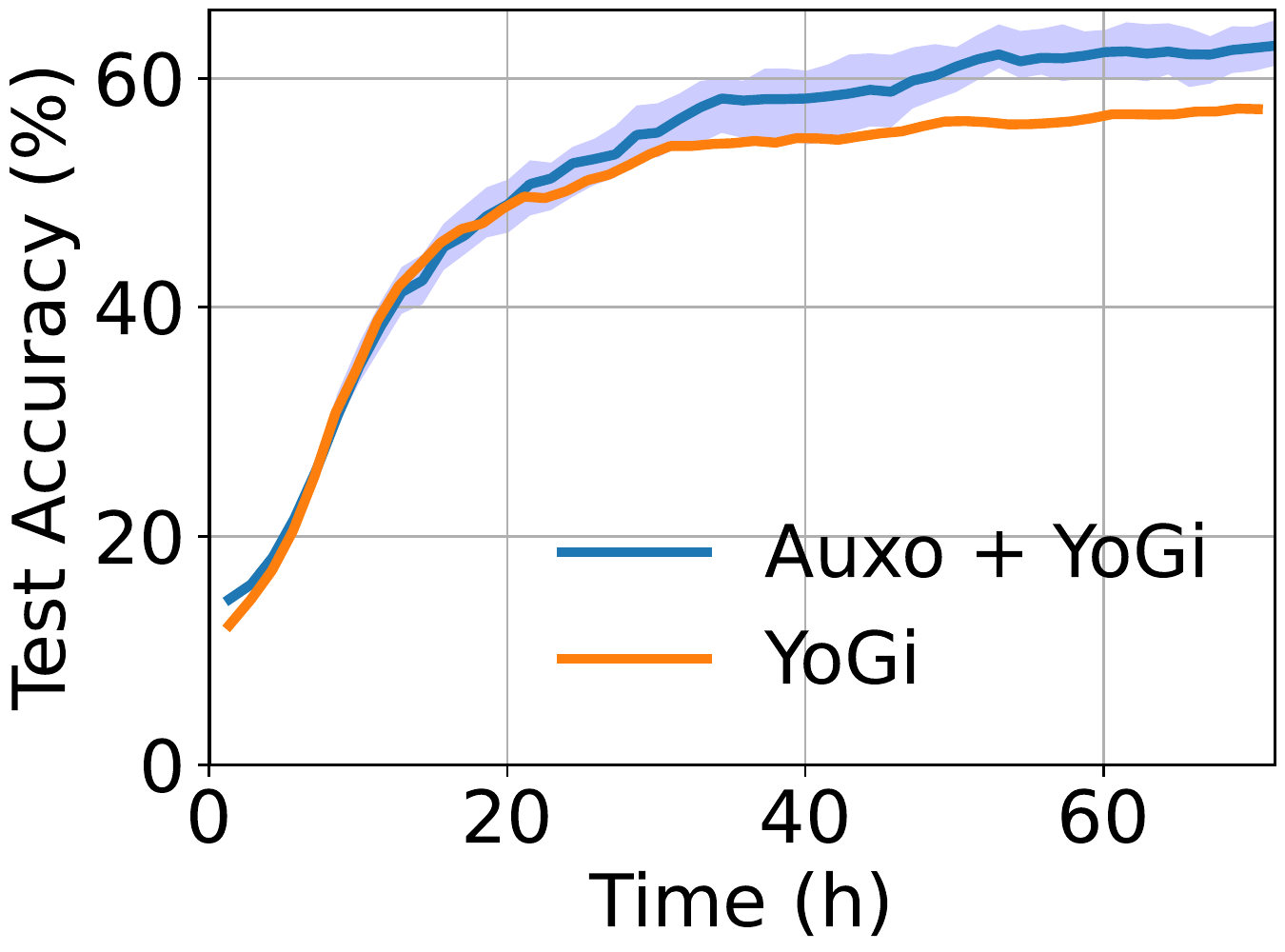}
    \caption{ShuffleNet (OpenImg) }%
    \label{fig:openimg1}
  \end{subfigure}
  \begin{subfigure}{1.6in}
    \centering
    \includegraphics[scale=0.3]{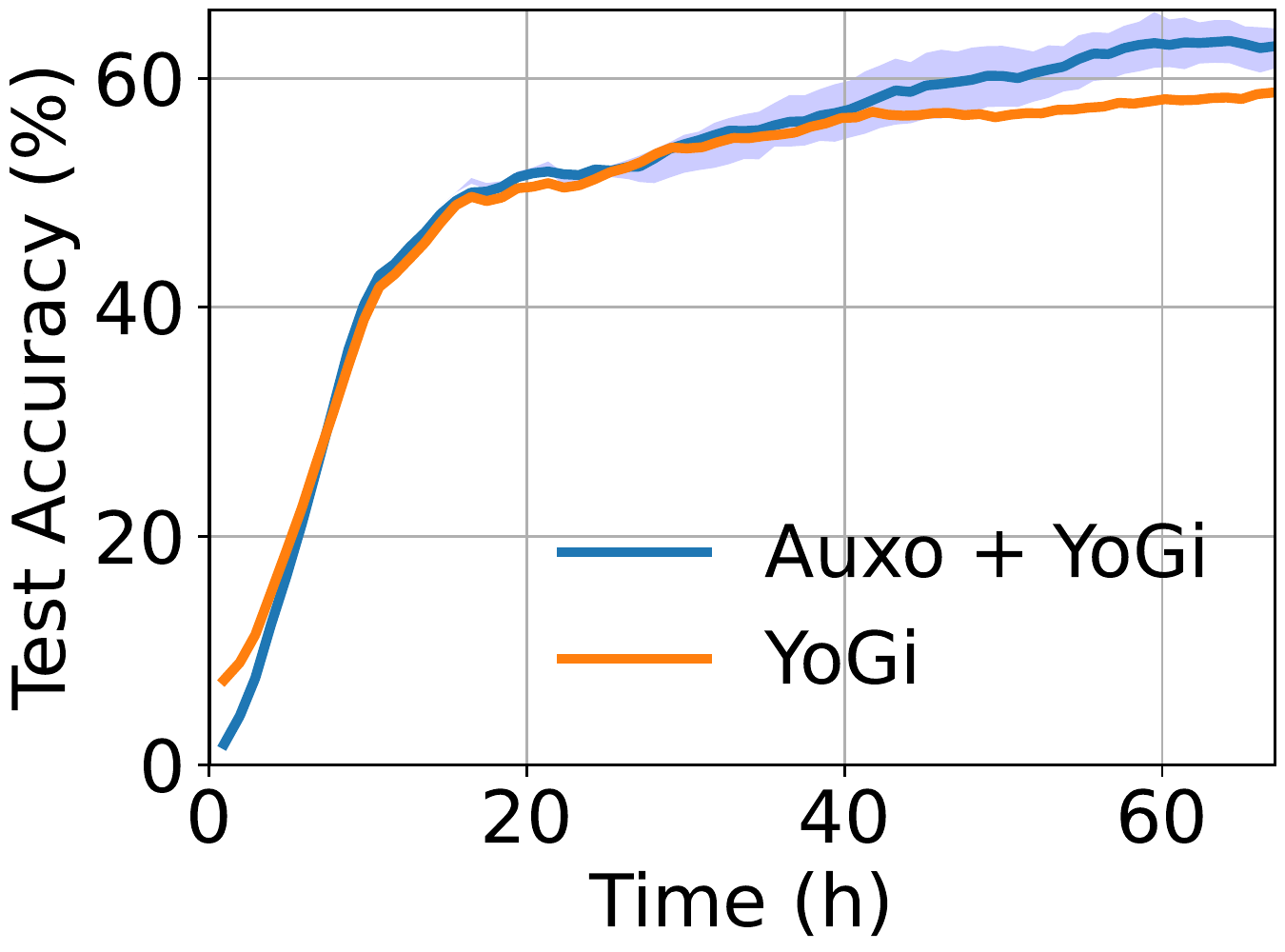}
    \caption{MobileNet (OpenImg)}%
   \label{fig:openimg2}
  \end{subfigure}
    \begin{subfigure}{1.62in}
    \centering
    \includegraphics[scale=0.3]{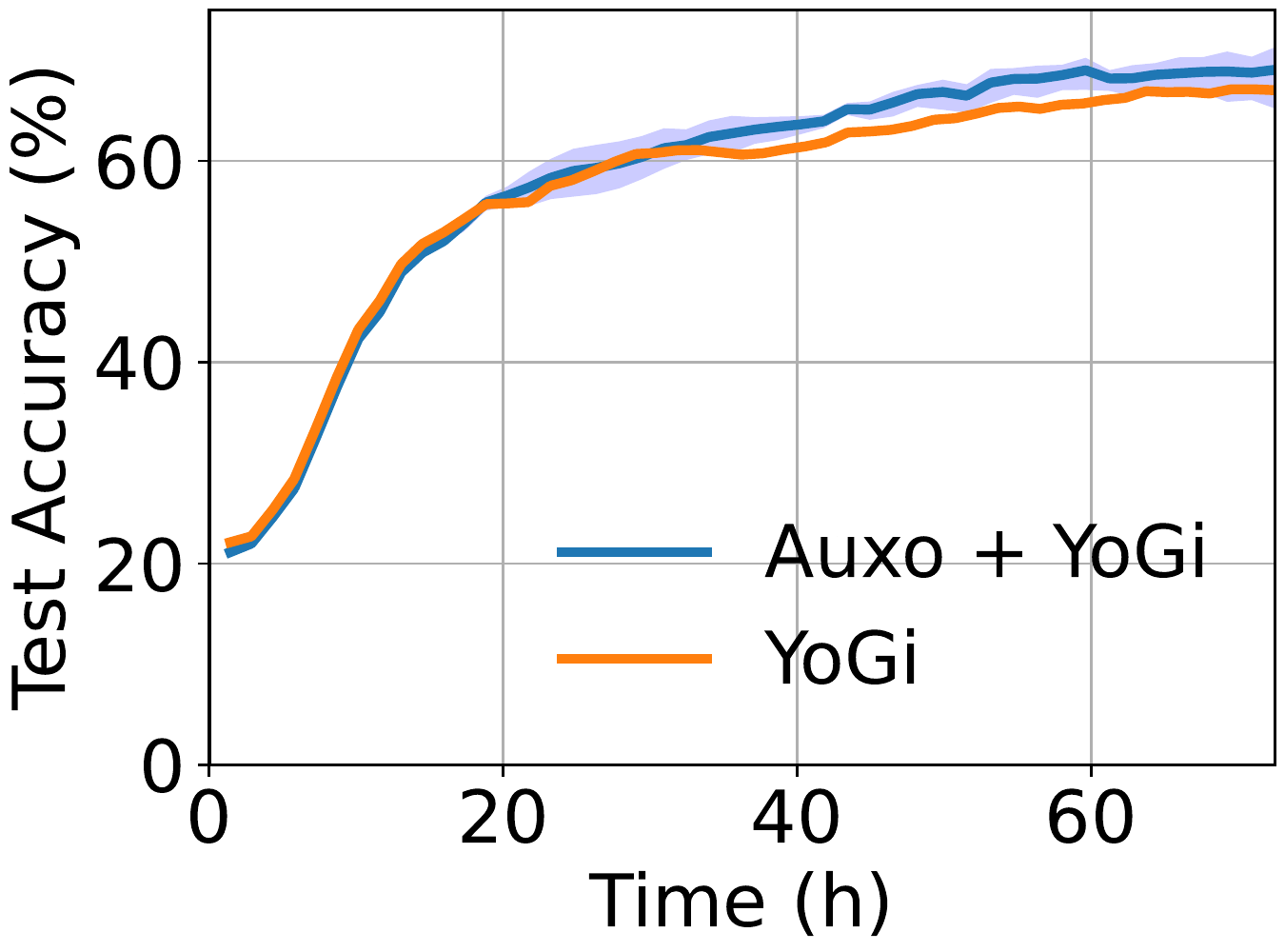}
    \caption{ShuffleNet (OpenImg-E)}%
    \label{fig:easy1}
  \end{subfigure}
    \begin{subfigure}{1.6in}
    \centering
    \includegraphics[scale=0.3]{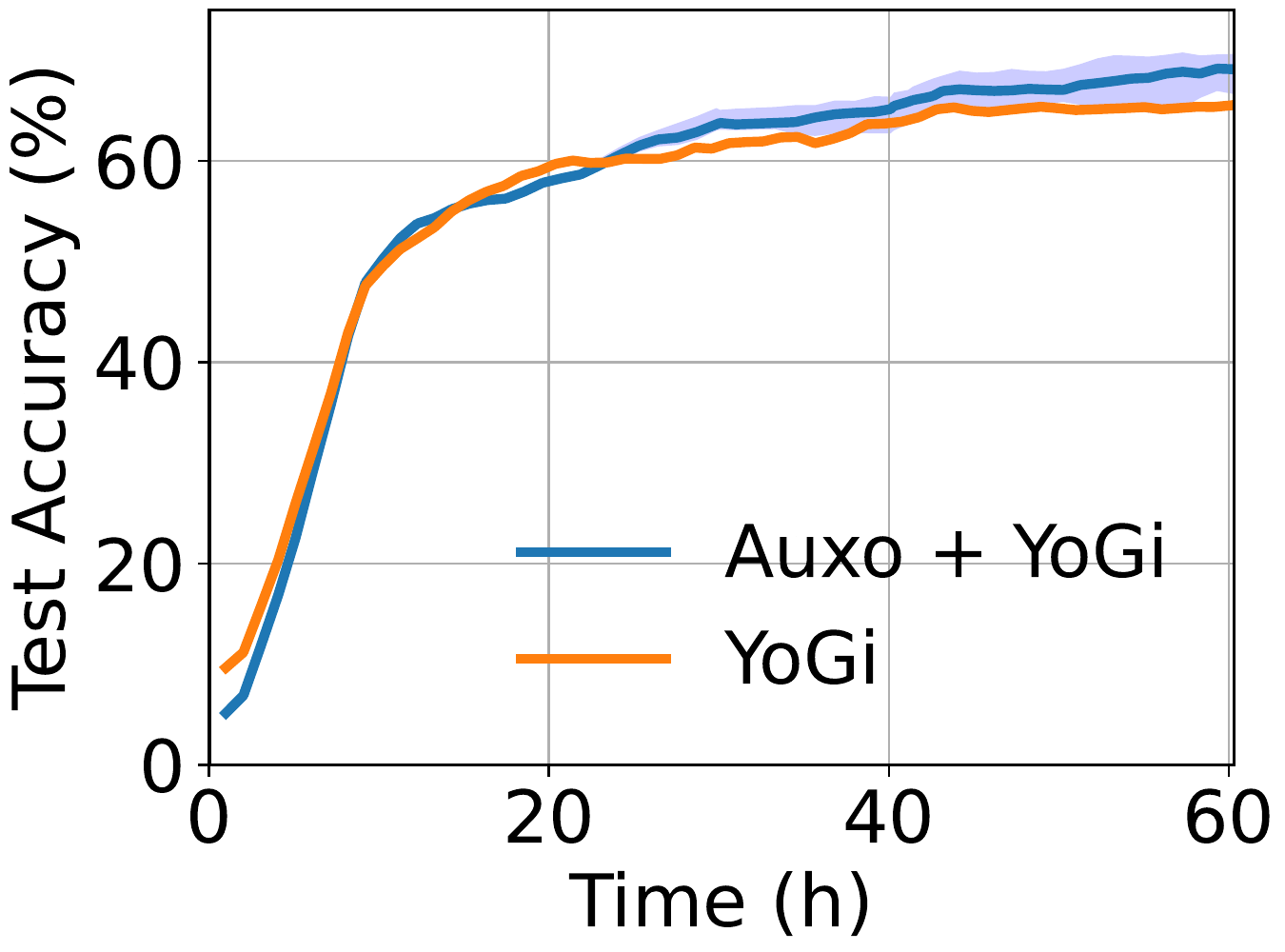}
    \caption{MobileNet (OpenImg-E)}%
    \label{fig:easy2}
  \end{subfigure}
  
 \begin{subfigure}{1.62in}
    \centering
    \includegraphics[scale=0.3]{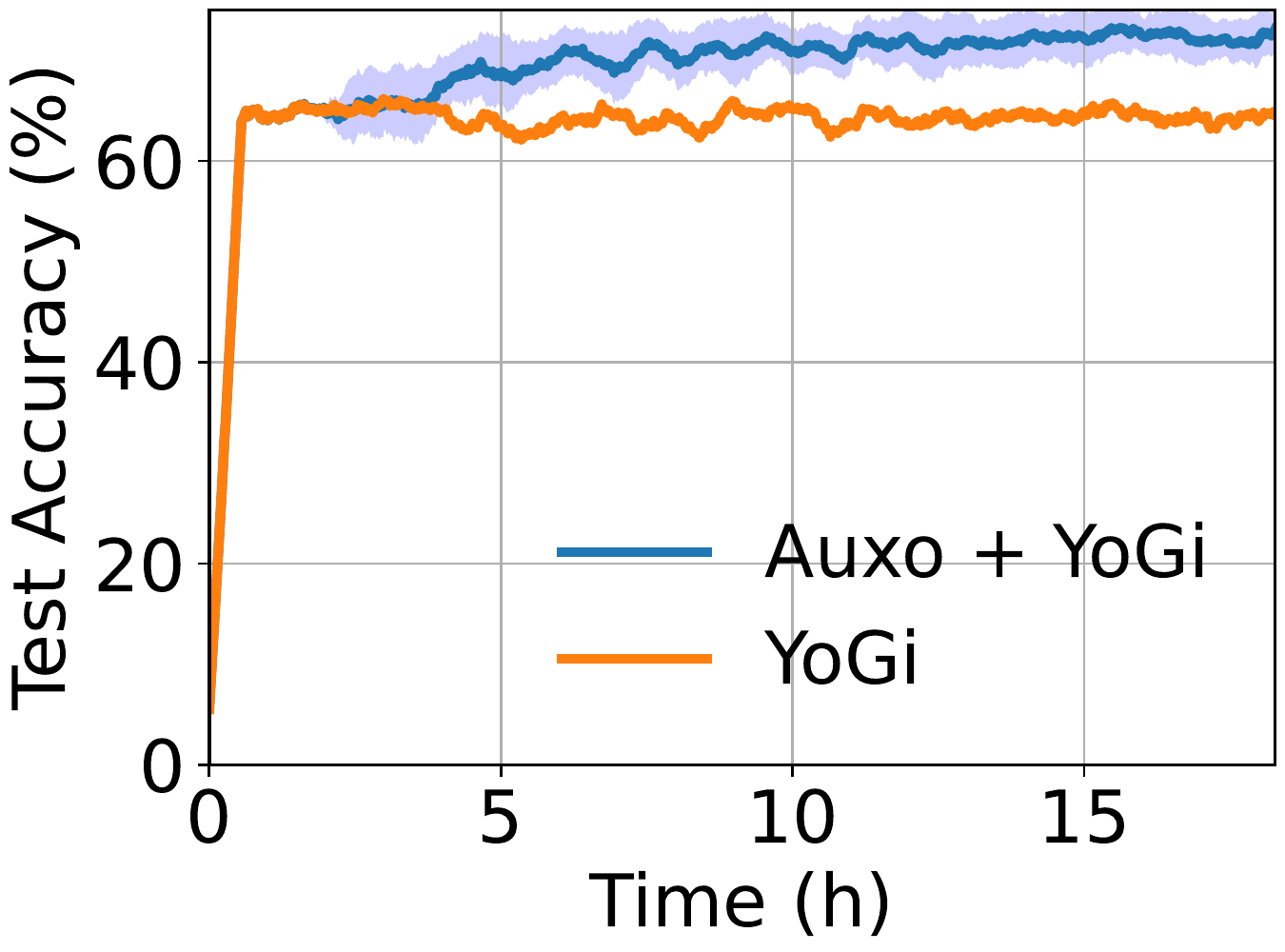}
    \caption{LR (Amazon) }%
    \label{fig:openimg}
  \end{subfigure}
  \begin{subfigure}{1.6in}
    \centering
    \includegraphics[scale=0.3]{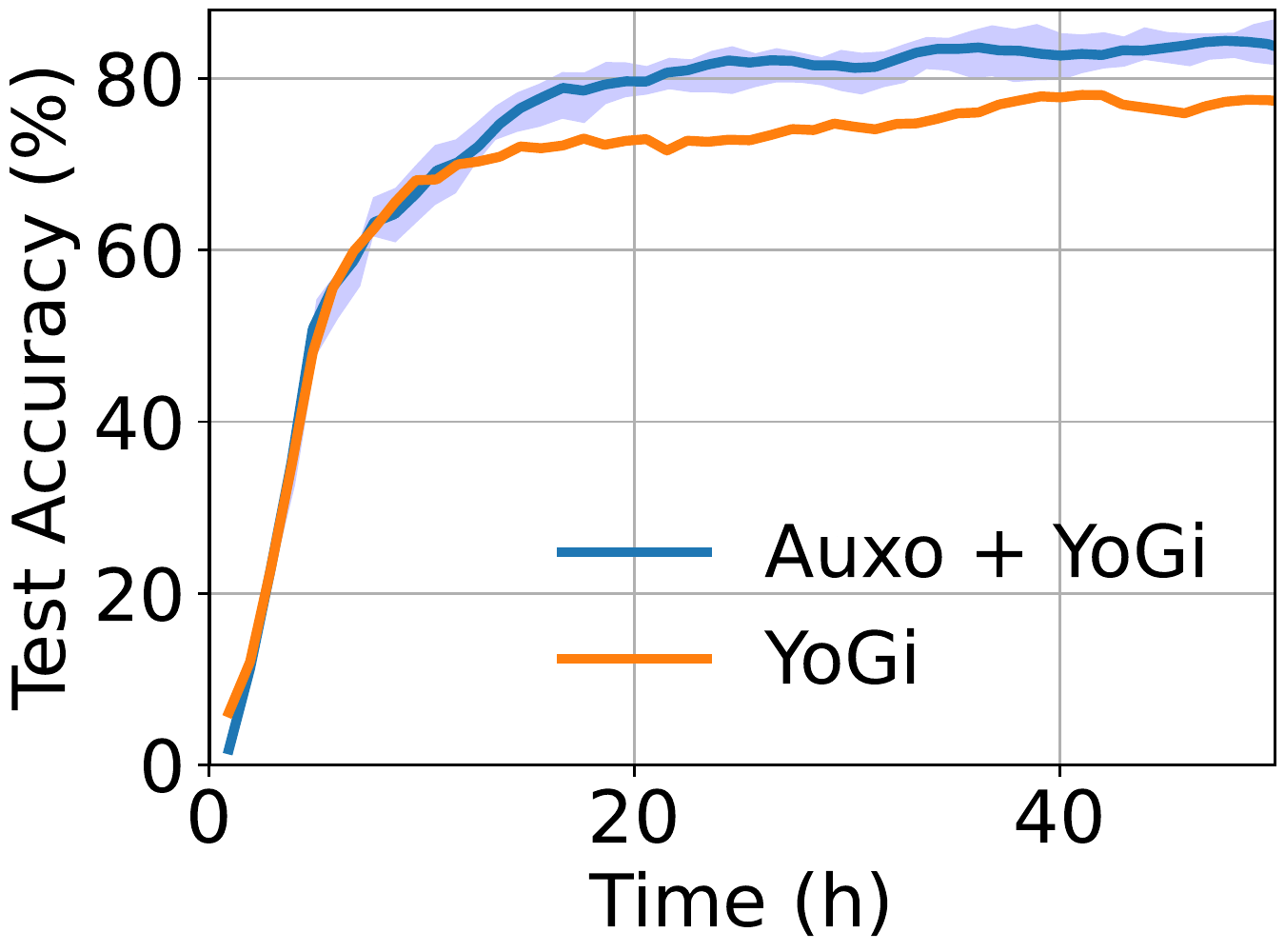}
    \caption{ResNet (FEMNIST)}%
   \label{fig:femnist}
  \end{subfigure}
    \begin{subfigure}{1.62in}
    \centering
    \includegraphics[scale=0.3]{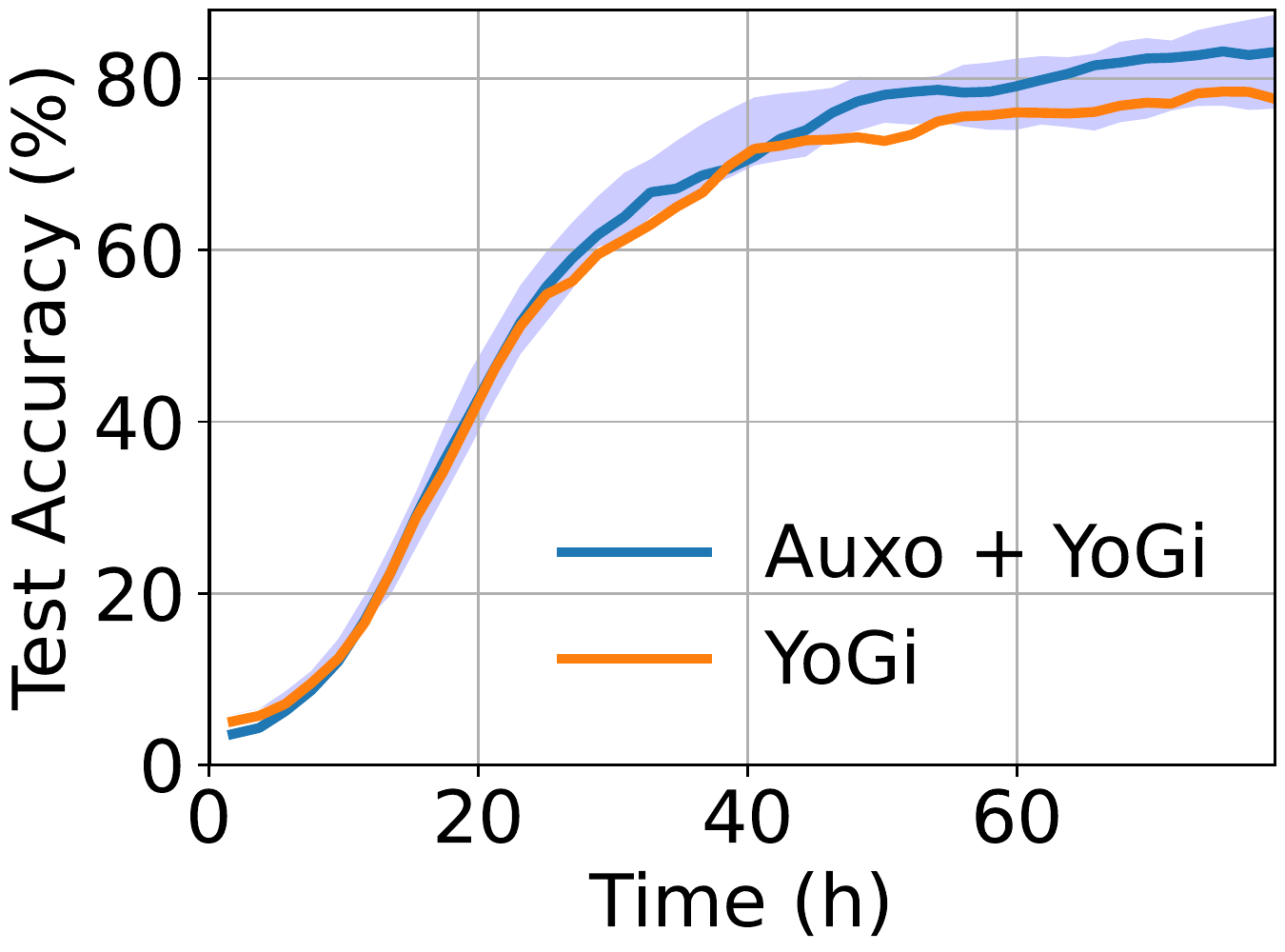}
    \caption{ResNet (Google Speech)}%
    \label{fig:speech}
  \end{subfigure}
    \begin{subfigure}{1.6in}
    \centering
    \includegraphics[scale=0.3]{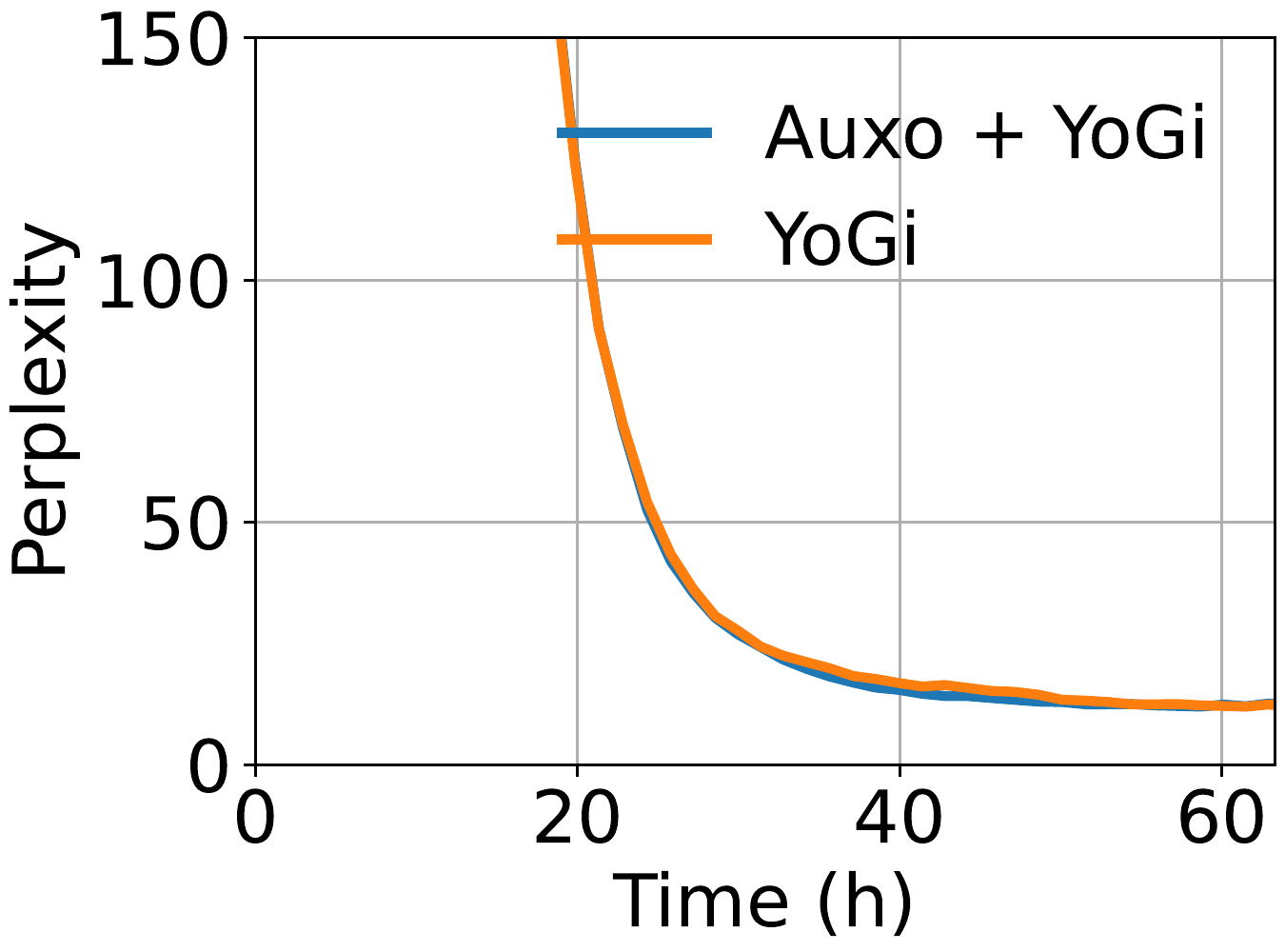}
    \caption{Albert (Reddit)}%
    \label{fig:reddit}
  \end{subfigure}
  \caption{Time-to-Accuracy performance on different dataset. 
  For the language modeling (LM) task, a lower perplexity is better.
  The solid line reflects the average test accuracy.
  The shaded portion covers the test accuracy performance among all cohorts generated by \name.}
     \label{fig:end-to-end}
  \end{figure*}
   
               \vspace{-.5em}
\subsection{Experiment Setup} 
 
\paragraph{Evaluation environment}
We use 24 NVIDIA Tesla P100 GPUs on CloudLab~\citep{cloudlab} to emulate the large-scale client training in our evaluations. 
The client data distribution follows the real-world partition, where client data can vary in quantities, data, and label distribution.
We use the open-source benchmark FedScale~\citep{fedscale} with standardized setup including realistic device capacity, data, and client availability traces. 
We report the simulated wall clock time by relying on these realistic FL system and data traces.

\paragraph{Datasets and models}

We run three categories of applications with six FL datasets~\citep{fedscale} of different scale factors using real-world partitions, whose statistics are reported in Table~\ref{table:data-stats}.
The clients for all datasets can check-in with \name multiple times following the availability trace.

\begin{enumerate}
\item \emph{Speech Recognition}: 
We train Resnet-34~\citep{resnet} on a small-scale Google Speech dataset with 35 commands.  
\item \emph{Image Classification}: 
We train Resnet-18 on small-scale FEMNIST with 62 handwritten digits to classify.
Also, we train ShuffleNet~\citep{shufflenet} and MobileNet~\citep{mobilenet} on middle-scale OpenImage with 596 classes to classify, whereas OpenImage-Easy only has 60 classes.
\item \emph{Language Modeling}: 
We train logistic regression (LR) on middle-scale Amazon Review for ratings prediction, and Albert model~\citep{albert} on large Reddit for word prediction.
\end{enumerate}

These applications are widely used in real end-device applications~\citep{mobile-app}, and these models are lightweight.
 \vspace{-.5em}

\paragraph{Parameters}
We follow the standardized experiment and parameter settings in FedScale. 
We adopt an over-commitment strategy to mitigate stragglers which allow 25\% failures or stragglers every round as in real FL deployments~\citep{tff-paper}.
We set the number of participants per round to be 200, the local minibatch size to be 6, and the initial learning rate  to be 4e-5 for the Albert model, and 0.05 for other models. 
And we use the linear scaling rule~\citep{lr} to scale the learning rate.

\paragraph{Metrics}
The \emph{time-to-accuracy} performance, \emph{final test accuracy}, and \emph{model bias} are our key metrics. 
We use the cohort member's test data, which follows the realistic data partition, to evaluate each cohort model.
The test data would be the global test data if we end up with one global model.
For each experiment, we report the average  top-1 accuracy based on the results over 3 runs.

\begin{figure*}[]
  \centering
 \begin{subfigure}{1.6in}
    \centering
    \includegraphics[scale=0.3]{openimg.pdf}
    \caption{Yogi}%
    \label{fig:yogi}
  \end{subfigure}
      \begin{subfigure}{1.6in}
    \centering
    \includegraphics[scale=0.3]{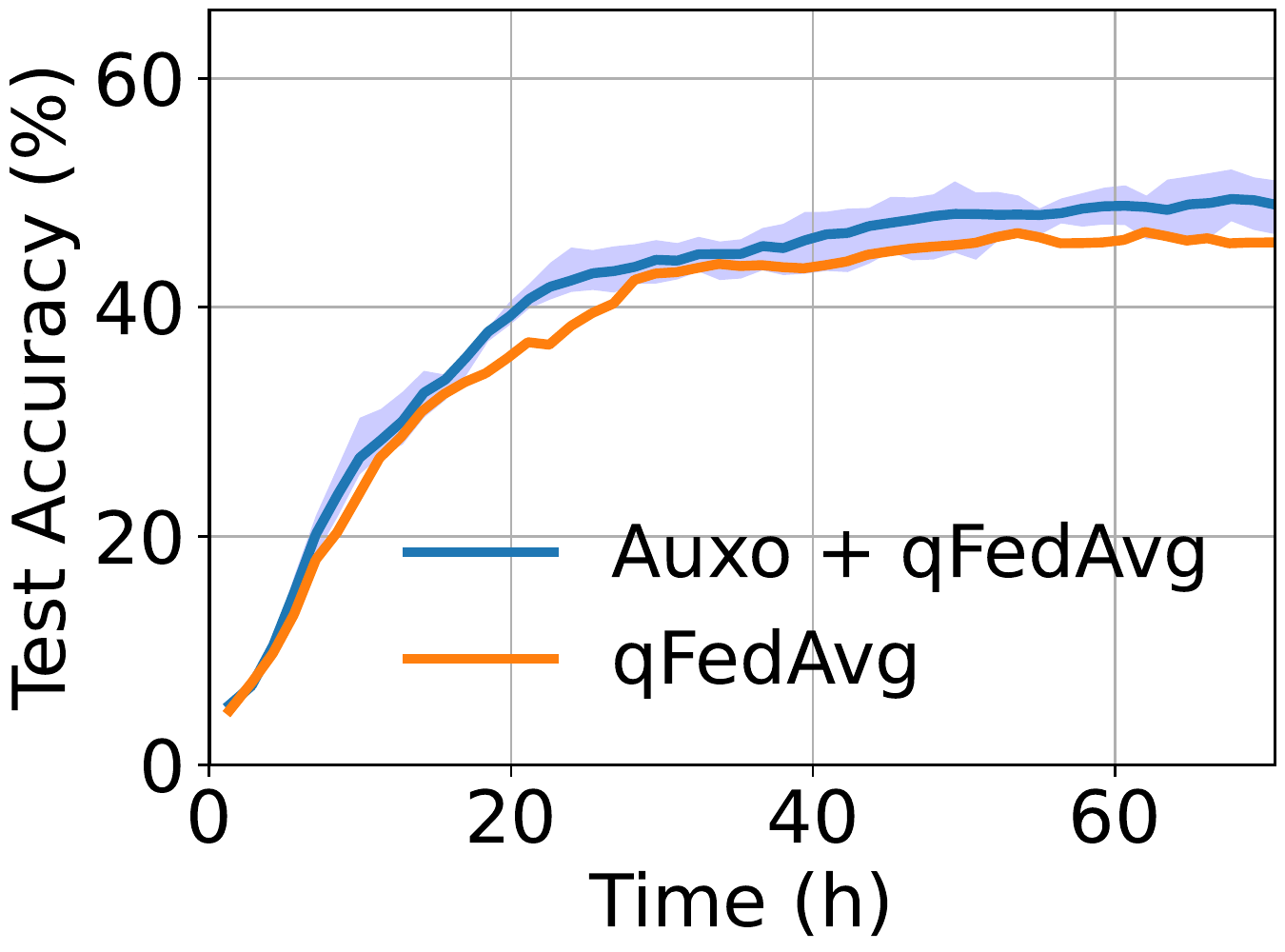}
    \caption{q-Fedavg}%
    \label{fig:qfedavg}
  \end{subfigure}
    \begin{subfigure}{1.6in}
    \centering
    \includegraphics[scale=0.3]{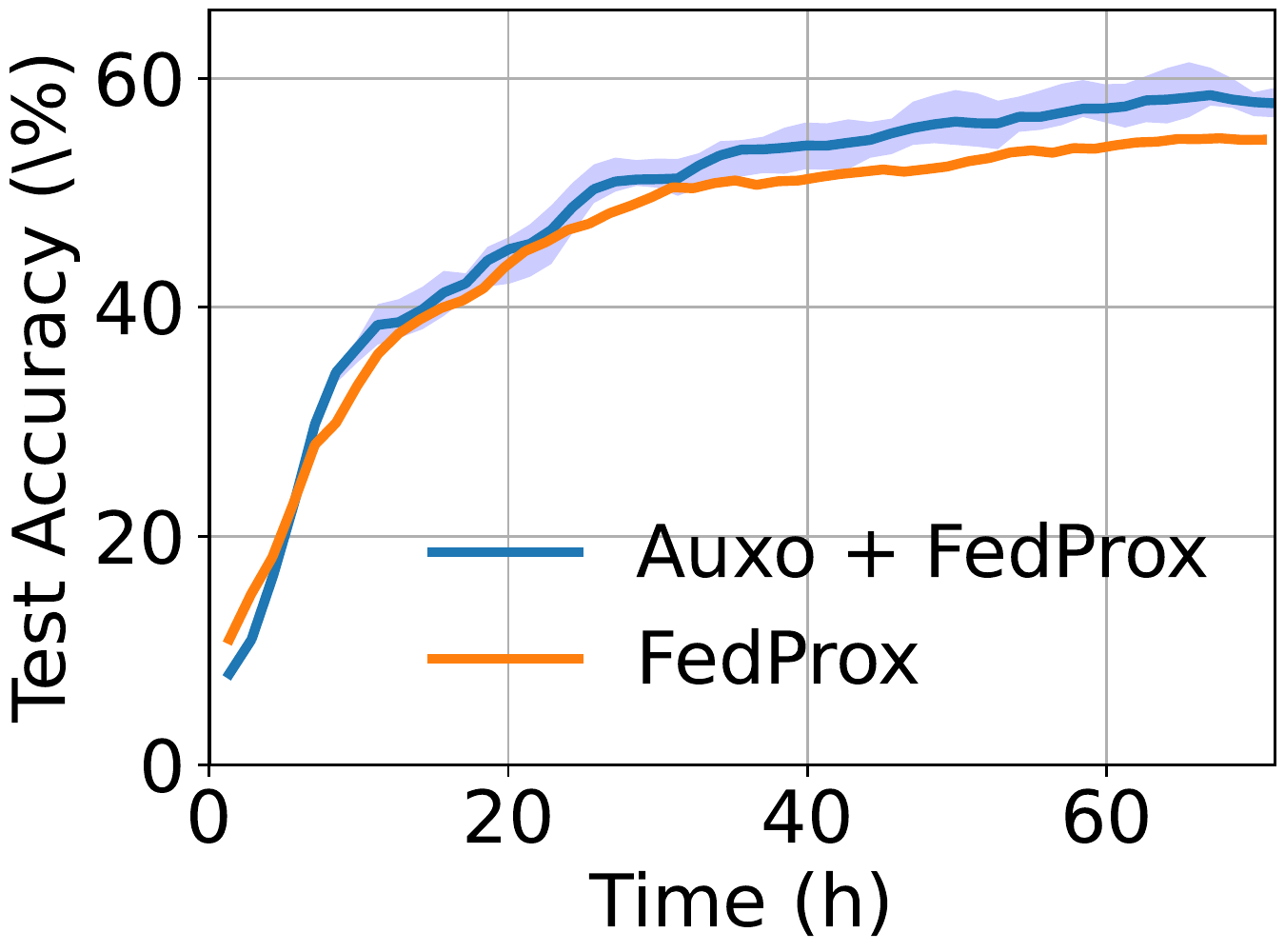}
    \caption{FedProx}%
    \label{fig:prox}
  \end{subfigure} 
    \begin{subfigure}{1.6in}
    \centering
    \includegraphics[scale=0.3]{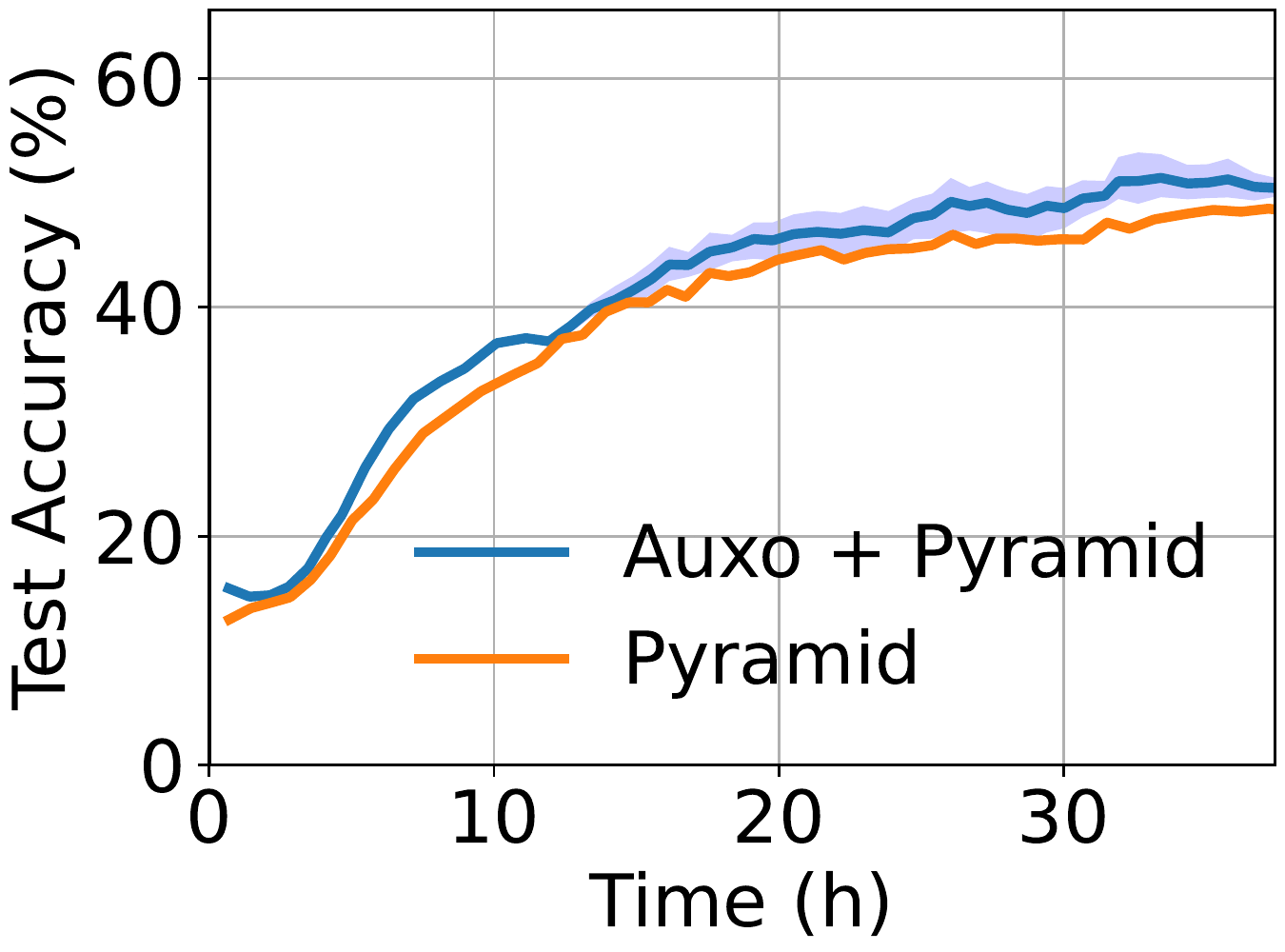} 
    \caption{PyramidFL + YoGi }
  \label{fig:pyramid}
  \end{subfigure} 
     \vspace{-0.5em}
  \caption{\name with different  FL algorithms. }
  \label{fig:diff-algo} 
   \end{figure*}
      
\subsection{End-to-End Performance}
\label{sec:e2e}
\paragraph{\name's performance on different datasets.}
We first evaluate \name's performance on different real-world FL datasets.   
In the following experiments, we adopt YoGi as the FL algorithm because it outperforms other FL algorithms most of the time.
Table~\ref{table:e2e-perf} summarizes the key time-to-accuracy performance of all datasets, where we tease apart the overall improvement with statistical and system ones.
We quantify the time-to-accuracy as speedup by \name, which measures how many times \name can speed up to achieve the target accuracy compared to the baseline time cost. 
Figure~\ref{fig:end-to-end} reports the timeline of training to achieve different accuracy, as different cohorts perform FL asynchronously. 
The shaded portion covers the test accuracy among all cohorts generated by \name.

We notice that \name speeds up the wall-clock time to reach target accuracy up to $2.2 \times $ faster.
Moreover, the final accuracy for different datasets is improved by 3.4\%--8.2\%.
The benefit of \name varies over datasets.
For most of the datasets, \name can achieve significant final accuracy improvement.
Nevertheless, \name does not improve Reddit task because the clients' texting behavior is similar to each other that makes it hard to identify significant groups as shown in Figure~\ref{fig:kmeans}.
Hence, \name decides not to partition into multiple cohorts to maximize the benefit of clustering in FL training.

 \paragraph{\name's performance on different FL algorithms.}
We then evaluate \name's performance on ShuffleNet-OpenImage with different FL Algorithms, which are complementary to \name. 
We refer to YoGi running atop \name as YoGi+\name, and similarly for FedProx, q-FedAvg and PyramidFL+YoGi.

As shown in Figure~\ref{fig:diff-algo},
 \name speeds up the time to reach the target accuracy of baseline algorithms, from $1.2\times$ to $2.2 \times$ faster and improve the final test accuracy by 3\%--6.8\%.

As for the personalization algorithm FTFA, we adopt the cohort models generated by \name to conduct local training using FTFA on corresponding cohort members. 
In addition to faster convergence of the initial model, \name also improves the average test accuracy of FTFA from 63.18\% to 67.40\% with local fine-tuning.

  
\paragraph{\name's benefit on resource efficiency.}
We finally show that \name can optimize resource efficiency on OpenImg dataset by saving 55\% training resources.
We also account for the affinity maintaining overhead into the client resource usage, which is around 0.02\% of the total resource consumption.

 \paragraph{\name's benefit on model bias.}
 
We show that \name can also mitigate the model bias due to smaller intra-cohort statistical heterogeneity.
We report the variance of the final accuracy distributions, the worst and best 10\% test accuracy in  Table~\ref{table:bias}.
Our experiment show that the variance of test accuracy is decreased for all the datasets by 4.8\% and 53.8\%.

 \subsection{Clustered FL Comparison}
\label{sec:exp-cfl}
We compare \name with four existing clustered FL algorithms CFL, FL+HC, FlexCFL, and IFCA in terms of three metrics: time-to-accuracy, resource-to-accuracy, and final accuracy.
Since these algorithms do not meet some real-world FL constraints (Table~\ref{table:comparison}), we simplify the settings accordingly.

We compare with CFL  in small-scale settings ($\sim 100$ clients) from the FEMNIST dataset to meet their full participation assumption. 
We observe little difference between CFL, \name, and baseline (\ie, no cohorts) in terms of time and resources used due to the absence of significant clusters within small populations.
However, this highlights the need for large-scale FL settings, where CFL cannot even be applied as it does not support partial participation.
 
\begin{figure}[t]       
  \centering
        \begin{subfigure}{0.22\textwidth}
            \centering
            \includegraphics[width=\textwidth]{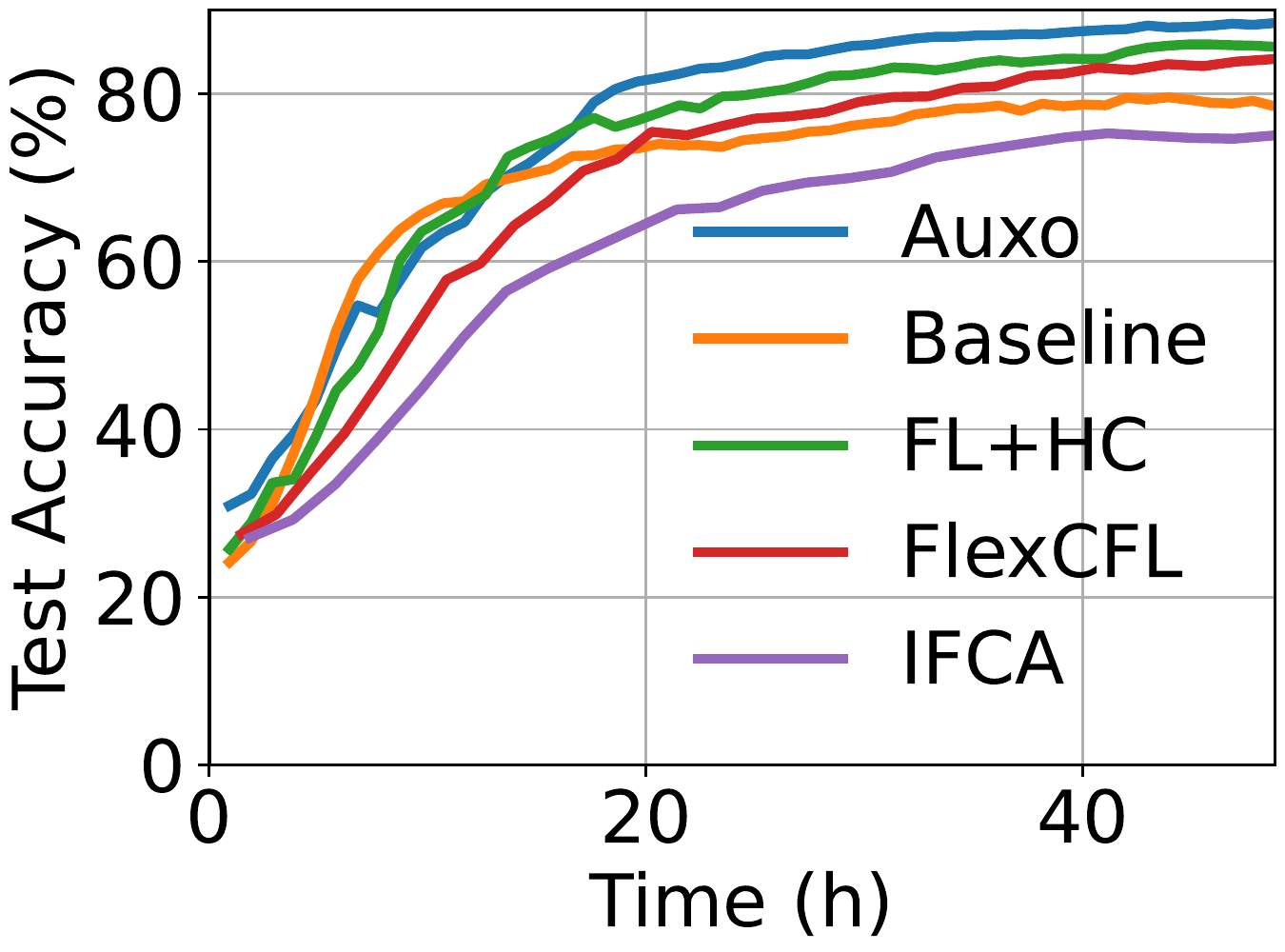} 
                 \caption{Accuracy-Time } 
         \end{subfigure}
         \hfill
        \begin{subfigure}{0.22\textwidth}
            \centering
            \includegraphics[width=\textwidth]{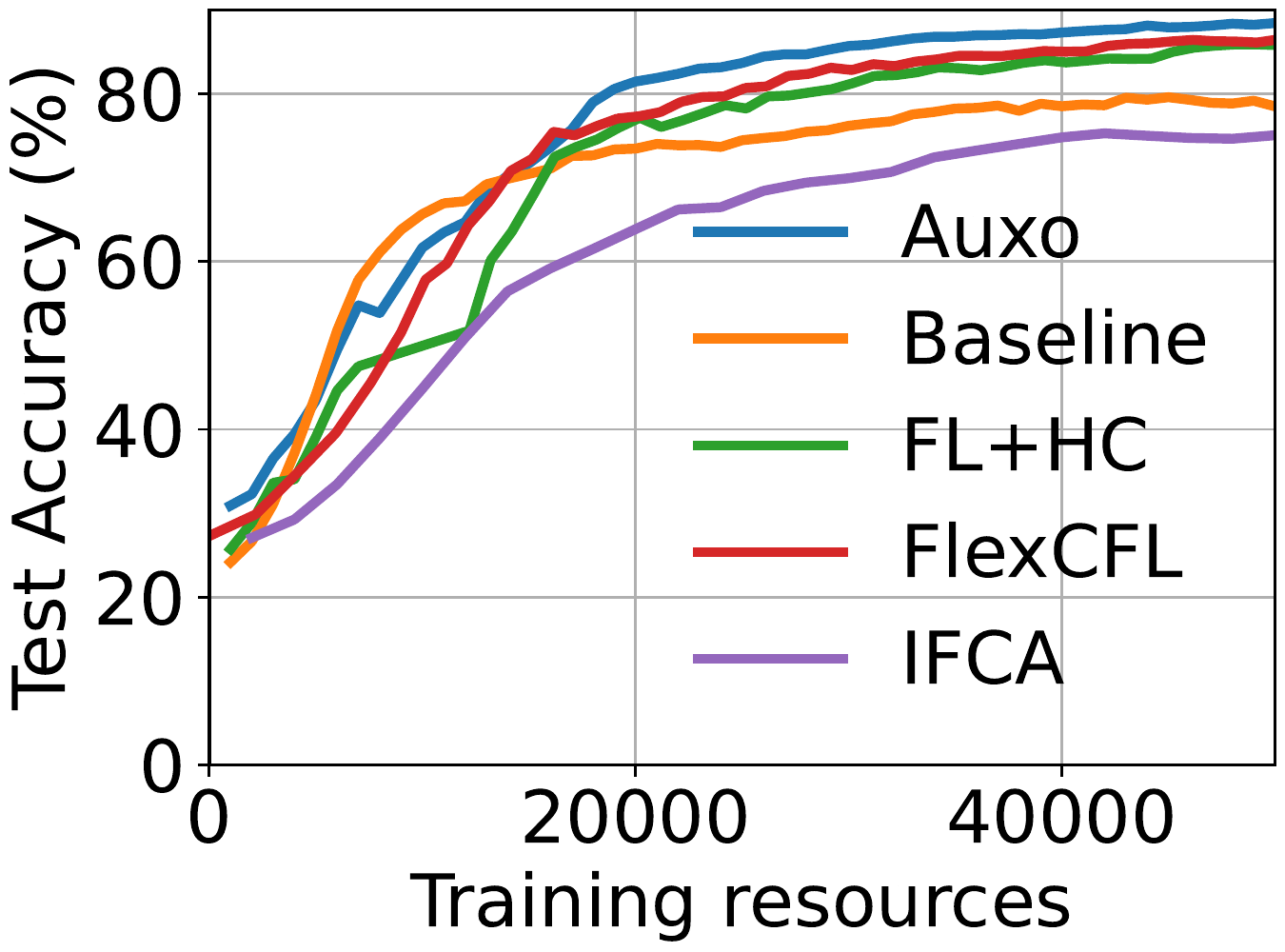} 
             \caption{Accuracy-Resource}  
        \end{subfigure}
    \caption{Comparison with clustered FL (FEMNIST). }
 \label{fig:addcfl}  
\end{figure}

To compare with FL+HC, FlexCFL and IFCA, we conduct experiments with the full FEMNIST and Amazon Review datasets \emph{without} the client availability traces to align with their constraints.
As shown in Table~\ref{table:relatedexp} and Figure~\ref{fig:addcfl},
 \name achieve better time efficiency $1.4\times-4.8\times$ and better resource efficiency  $1.3\times-4.8\times$  compared to the related works especially for the large-scale Amazon dataset, due to our efficient and scalable algorithm design. 
Also, our result for IFCA are consistent with  Motley~\citep{motley}.

\begin{table}[]
\resizebox{\columnwidth}{!}{
\begin{tabular}{c|c|ccc}
\hline
Dataset               & \multicolumn{1}{l|}{Setting} & \begin{tabular}[c]{@{}c@{}}Worst 10\% \\ (\%)\end{tabular} & \begin{tabular}[c]{@{}c@{}}Best 10\% \\ (\%)\end{tabular} & Variance   \\ \hline
\rowcolor{Gray}
OpenImg               & \name                  & 38                                 & 83                                        & 267   \\
\multicolumn{1}{l|}{} & Baseline              & 34                                      & 79                                               & 296   \\ \hline
\rowcolor{Gray}
OpenImg               & \name                  & 38                                    & 88                         & 234    \\
-Easy                 & Baseline              & 33                                           & 84                                    & 273    \\ \hline
\rowcolor{Gray}
Review                & \name                  & 50                                         & 100                                     & 460   \\
\multicolumn{1}{l|}{} & Baseline              & 32                               & 100                                      & 995   \\ \hline
\rowcolor{Gray}
FEMNIST               & \name                  & 63                                        & 97                                      & 171   \\
\multicolumn{1}{l|}{} & Baseline              & 60                                         & 93                                              & 185   \\ \hline
\rowcolor{Gray}
Speech                & \name                  & 57                                   & 100                                      & 479  \\
\multicolumn{1}{l|}{} & Baseline              & 52                                        & 100                                          & 503  \\ \hline
\end{tabular}
}
\caption{Summary of improvements on model bias.} 
               \vspace{-1.5em}
\label{table:bias}
\end{table}
 
 \begin{table}[] 
\centering
\resizebox{\columnwidth}{!}{
\begin{tabular}{lc|
>{\columncolor[HTML]{FFFFFF}}c ccc}
\hline
                                               & \multicolumn{1}{l|}{}                  & {\color[HTML]{000000} FL+HC}                               & FlexCFL                             & IFCA                                & \name                \\ \hline
\multicolumn{1}{l|}{}                          & \cellcolor[HTML]{EFEFEF}Speedup      & \cellcolor[HTML]{EFEFEF}{\color[HTML]{000000} $1.7\times$} & \cellcolor[HTML]{EFEFEF}$1.3\times$ & \cellcolor[HTML]{EFEFEF}$0.5\times$ & \cellcolor[HTML]{EFEFEF}$2.4\times$ \\
\multicolumn{1}{l|}{}                          & Efficiency                          & {\color[HTML]{000000} $1.6\times$}                         & \cellcolor[HTML]{FFFFFF}$1.8\times$ & $0.5\times$                         & \cellcolor[HTML]{FFFFFF}$2.4\times$ \\
\multicolumn{1}{l|}{\multirow{-3}{*}{FEMNIST}} & \cellcolor[HTML]{EFEFEF}Final acc. & \cellcolor[HTML]{EFEFEF}{\color[HTML]{000000} 5.8\%}      & \cellcolor[HTML]{EFEFEF}7.1\%       & \cellcolor[HTML]{EFEFEF}1.3\%       & \cellcolor[HTML]{EFEFEF}9.1\%       \\ \hline
\multicolumn{1}{l|}{~Amazon}                    & Speedup                              & {\color[HTML]{000000} $0.4\times$}                         & $ 0.4\times$                         & \cellcolor[HTML]{FFFFFF}$0.5\times$ & $2.3\times$                         \\
\multicolumn{1}{l|}{~ Review}                    & \cellcolor[HTML]{EFEFEF}Efficiency  & \cellcolor[HTML]{EFEFEF}{\color[HTML]{000000} $0.4\times$} & \cellcolor[HTML]{EFEFEF}$0.5\times$ & \cellcolor[HTML]{EFEFEF}$0.5\times$ & \cellcolor[HTML]{EFEFEF}$2.1\times$ \\
\multicolumn{1}{l|}{}                          & Final acc.                          & {\color[HTML]{000000} -2.9\%}                              & -2.7\%                               & 0.6\%                               & 5.4\%                               \\ \hline
\end{tabular}
}
\caption{Summary of improvements over baseline (\ie, no cohorts) in terms of time, resource and accuracy.  } 
               \vspace{-1.5em}
\label{table:relatedexp} 
\end{table}

\subsection{Sensitivity Analysis }
\label{sec:partition-time}
\begin{figure*}[t]
\begin{minipage}[t]{0.94\textwidth}
   \centering
    \begin{subfigure}[t]{0.33\textwidth}
   \centering
     \includegraphics[trim=0 0 0 0,clip,width=0.79\textwidth]{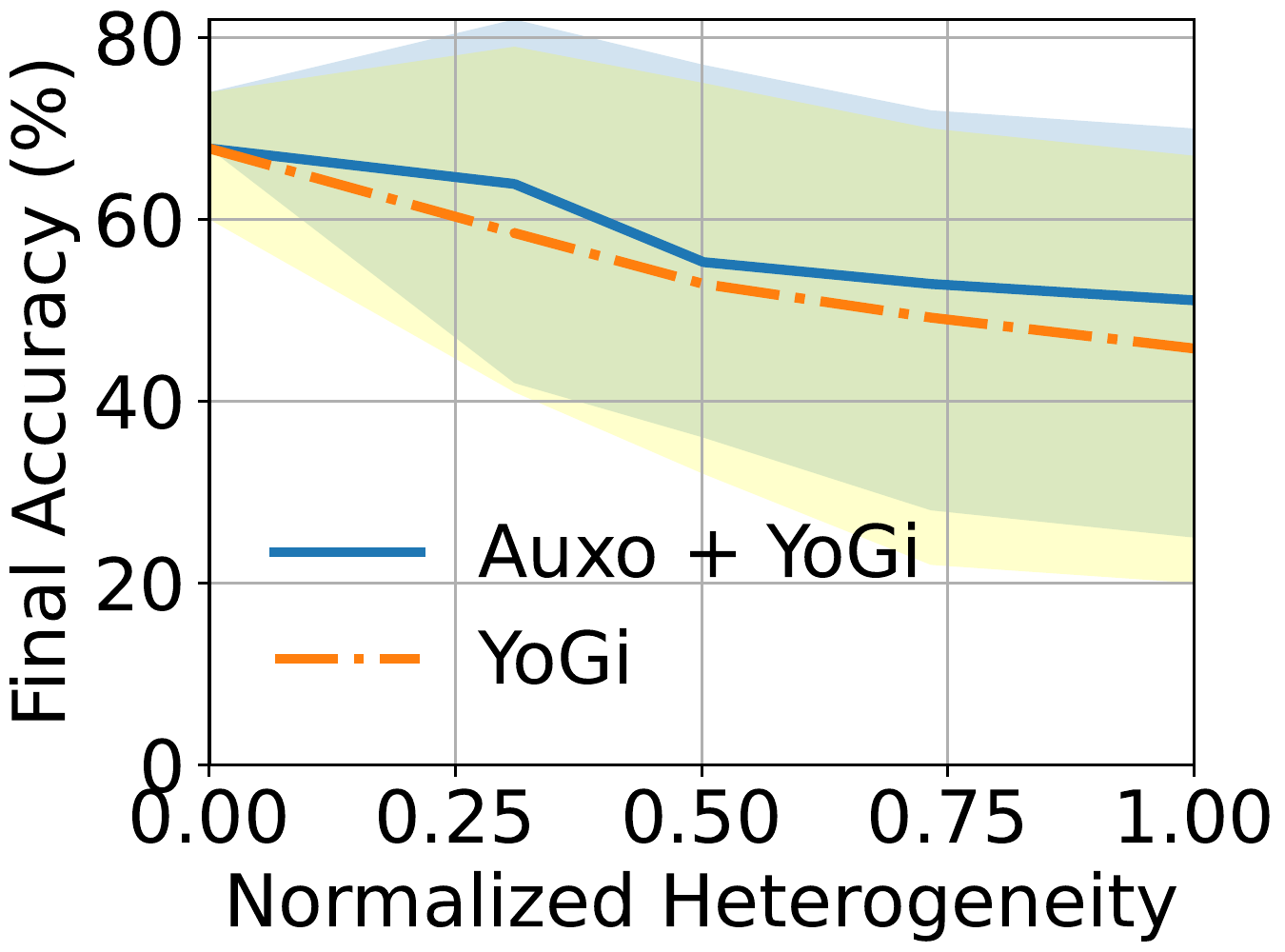}
   \caption{Different degrees of heterogeneity.}
    \label{fig:diff-heter}   
  \end{subfigure} 
  \hfill
  \begin{subfigure}[t]{0.3\textwidth}
  \centering
      \includegraphics[width=0.8\textwidth]{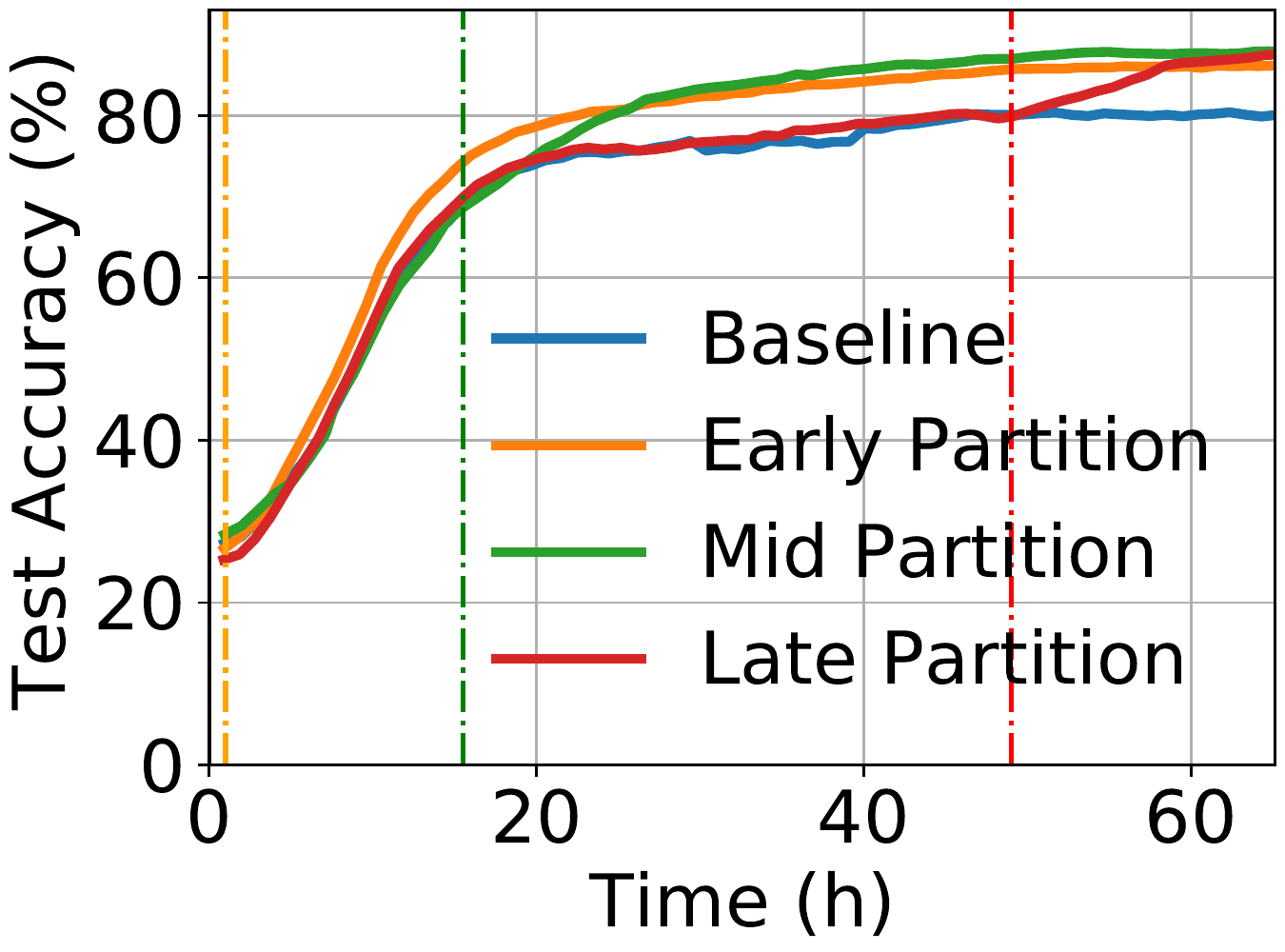}
     \caption{Different partition time.}
     \label{fig:conf} 
   \end{subfigure}
   \hfill
   \begin{subfigure}[t]{0.3\textwidth}
 \centering
    \includegraphics[width=0.8\textwidth]{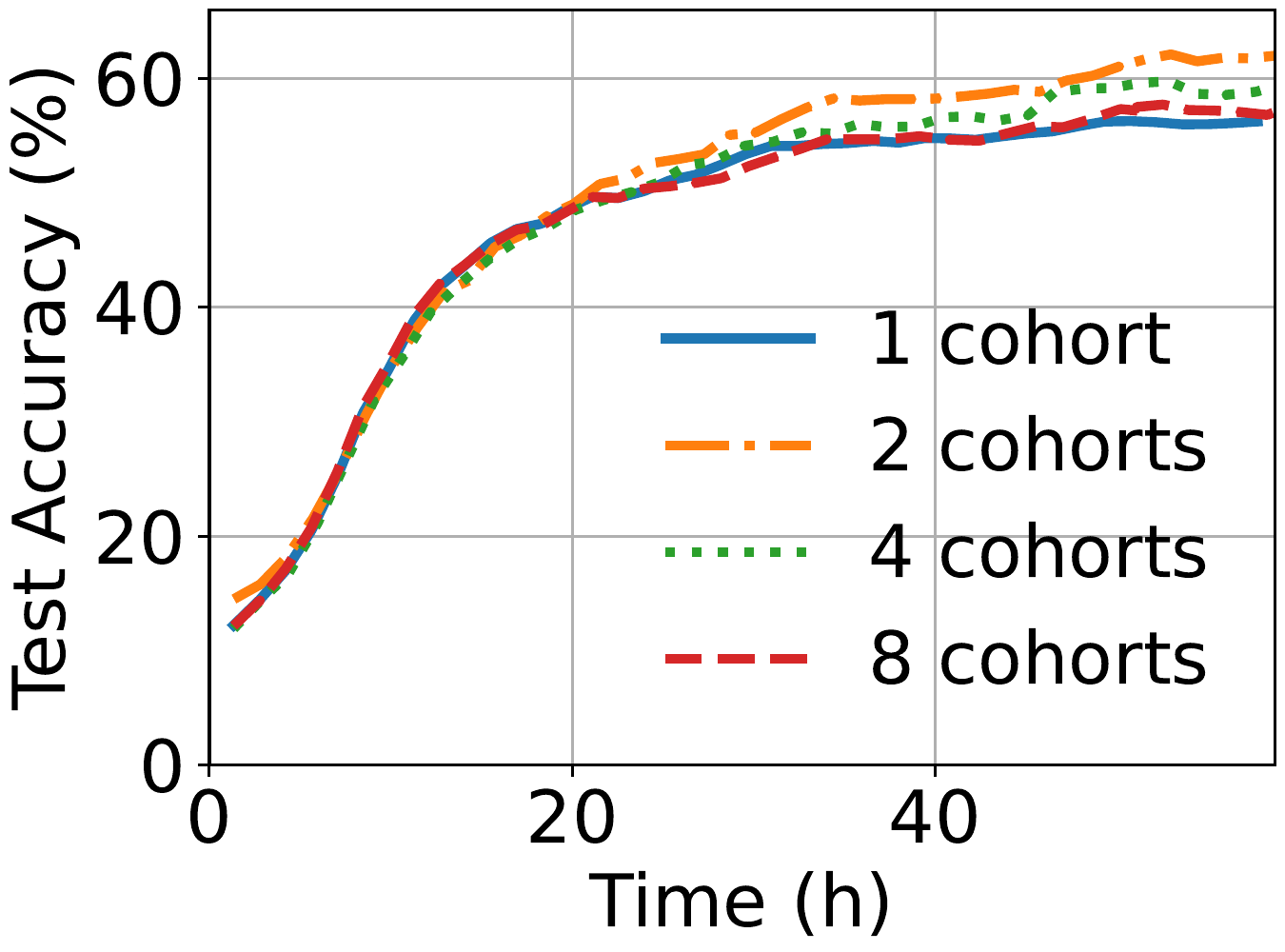}
    \caption{Different number of cohorts. }
    \label{fig:impact-cohort} 
  \end{subfigure}  
   \vspace{-0.5em}
      \caption{Sensitivity analysis. }  
         \end{minipage}
 \end{figure*}

\paragraph{Impact of different degrees of heterogeneity.}
We generate different statistical heterogeneity by applying affine shift~\citep{affine} on OpenImage, then we evaluate \name with YoGi across different degrees of statistical heterogeneity. 
Figure~\ref{fig:diff-heter} reports the final test accuracy as well as top 10\% and worst 10\% client test accuracy on different degrees of heterogeneity.
We observe that \name can improve model accuracy and mitigate model bias under different degrees of heterogeneity.
Moreover, similar to the previous experiment, \name achieves faster time-to-accuracy performance from  $1.2\times$ to $1.8 \times$.
    
 \paragraph{Impact of time to partition. } 
 We investigate the impact of different cohort partition times on the model convergence.
As mentioned in Section~\ref{sec:system-tradeoff}, the partition time relates to the trade-off between model generalizability and intra-cohort heterogeneity.
As shown in Figure~\ref{fig:conf}, we choose different partition times with the same cohort composition and report the test accuracy to time performance on FEMNIST. 
We observe that cohort-based training all outperform the baseline experiment with one cohort.
However, early partitions such as FlexCFL and IFCA are worse than intermediate partitions, because it sacrifices the model generalizability. 
Similarly, late partition after convergence as CFL does not outperform intermediate partition, because it slows down the model convergence to a smaller heterogeneous population.

\begin{figure}[]
        \centering 
            \centering
            \includegraphics[width=0.25\textwidth]{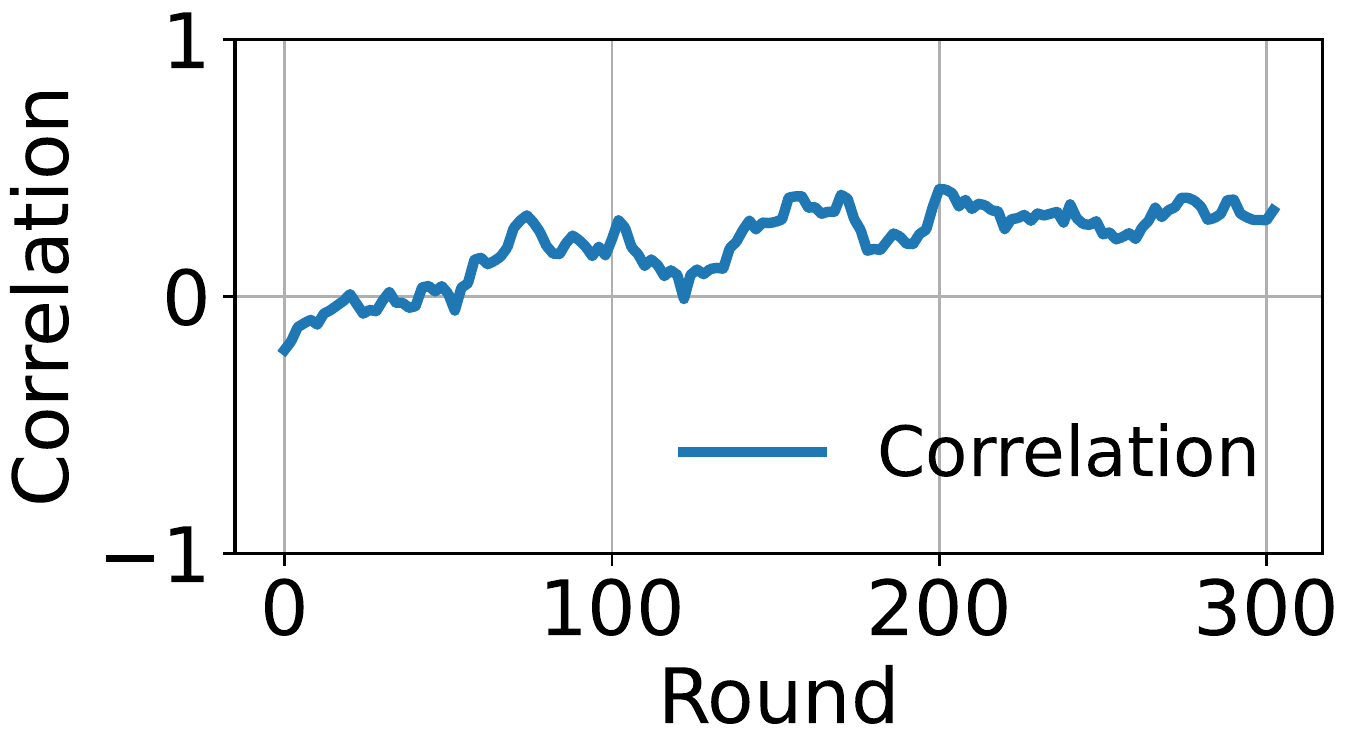}  
               \vspace{-0.5em}
    \caption{Impact of cluster start time. }
                   \vspace{-1.5em}

  \label{fig:cluster-start} 
\end{figure}

 \paragraph{Impact of time to start clustering. }  
 We investigate the impact of  the clustering start time on the model convergence.
As mentioned in Section~\ref{sec:system-tradeoff}, different clustering start time may affect the clustering accuracy and efficiency. 
To quantify how well the gradient similarity correlates with data similarity, we use Pearson correlation coefficient $   r = \frac{ \sum_{i=1}^{n} (G_i - \bar{G})(D_i - \bar{D}) }{ \sqrt{\sum_{i=1}^{n} (G_i - \bar{G})^2 \sum_{i=1}^{n} (D_i - \bar{D})^2} }$, where D and G are pairwise data similarity and gradient similarity.
As shown in Figure~\ref{fig:cluster-start}, the similarity correlation slightly increases over training rounds, which suggests a slightly later cluster start time.

\paragraph{Impact of the number of cohorts.}
We investigate the impact of the number of cohorts generated by \name, which relates to the trade-off between training resources and intra-cohort heterogeneity (\S\ref{sec:system-tradeoff}).
As shown in Figure~\ref{fig:impact-cohort}, we observe that the model convergence is negatively affected once the number of cohorts exceeds 4 under the same resource budget.
By further comparing the reduce of heterogeneity with different number of cohorts indicated in Figure~\ref{fig:kmeans}, 
this result verifies that better model convergence can be achieved as long as the heterogeneity can proportionally compensate the reduced training resources
 
\begin{figure*}[t]
\begin{minipage}[t]{0.94\textwidth}
  \centering 
   \begin{subfigure}[t]{0.25\textwidth}
    \centering
    \includegraphics[trim=0 0 0 0,clip,width=\textwidth]{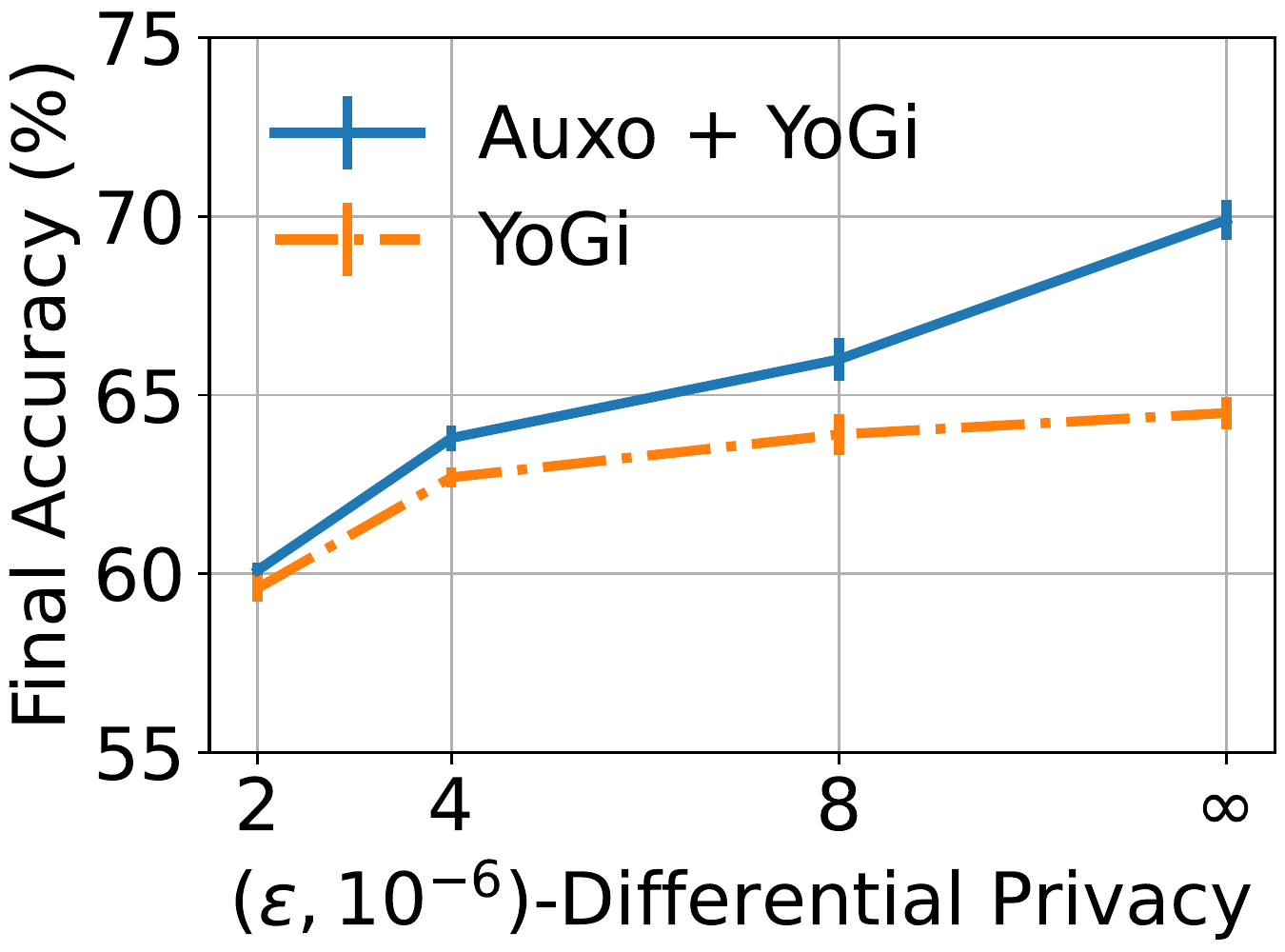}
     \caption{Local DP. }%
    \label{fig:dp}
  \end{subfigure}
  \hfill
  \begin{subfigure}[t]{0.25\textwidth}
    \centering
    \includegraphics[trim=0 0 0 0,clip,width=\textwidth]{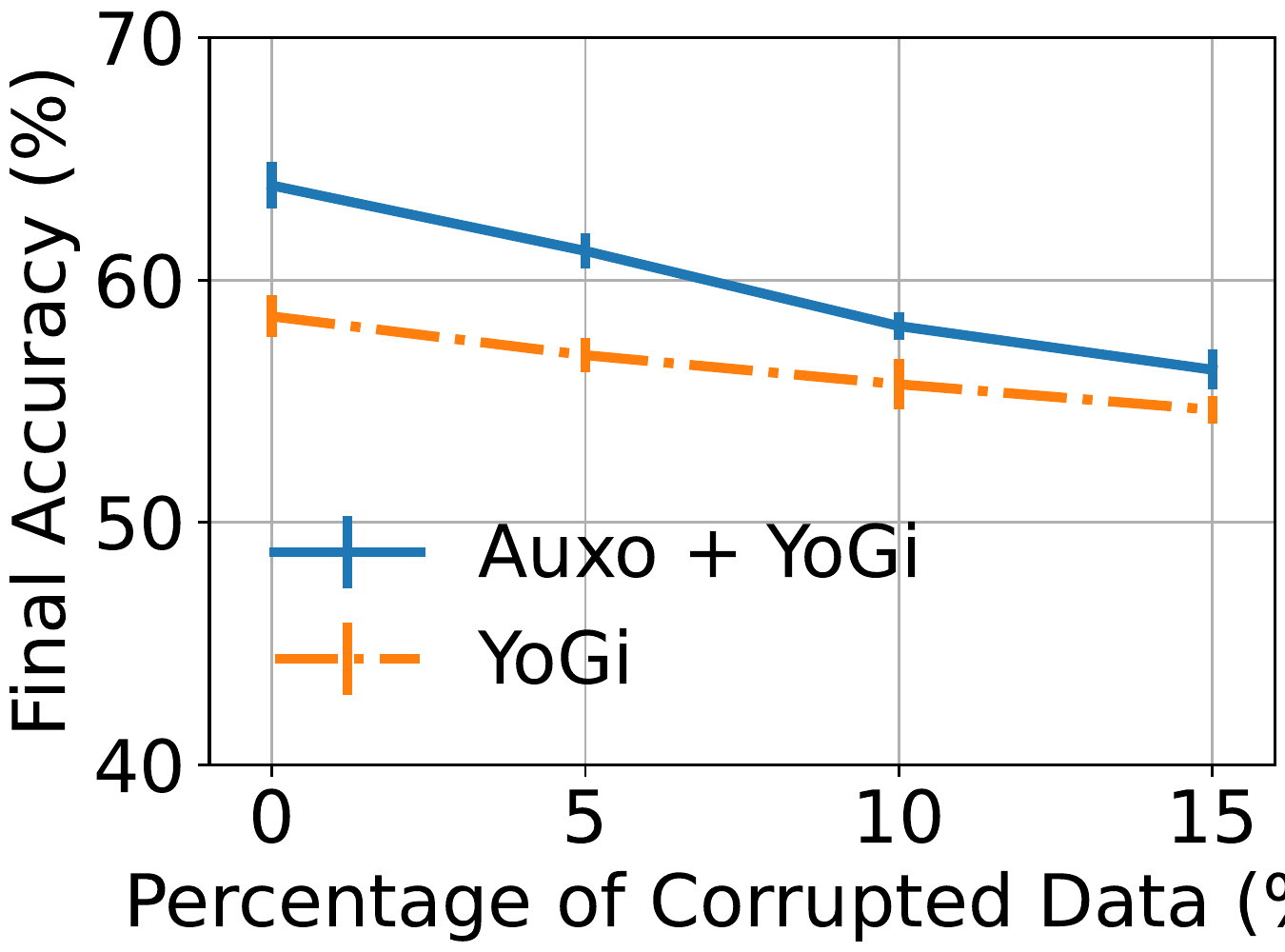}
     \caption{Corrupted client.}%
    \label{fig:malicious}
  \end{subfigure}
  \hfill
  \begin{subfigure}[t]{0.25\textwidth}
    \centering
    \includegraphics[width=\textwidth]{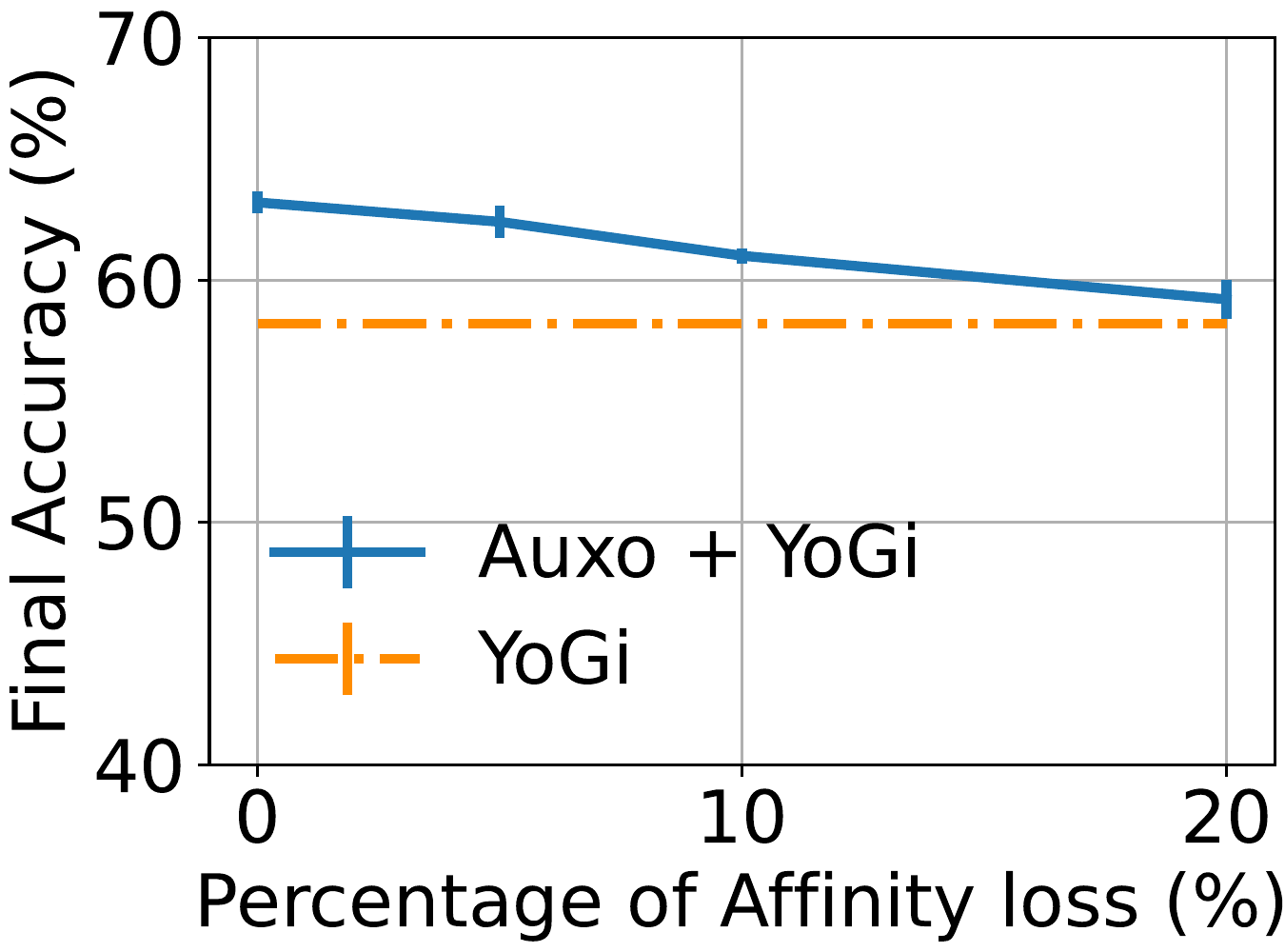}
     \caption{Client affinity loss.}
    \label{fig:client-fail}
  \end{subfigure}
     \vspace{-0.5em}
    \caption{Robustness of \name under different scenarios. }
   \end{minipage}
\end{figure*}

\subsection{\name  Resilience }
\label{sec:dp}
\paragraph{Impact of differential privacy.}
\name is robust to local differential privacy (LDP). 
LDP is used to protect user-level privacy by adding Gaussian noise to the client update before sending it to the server, but it hurts model accuracy.
We evaluate \name's performance under LDP for a learning task on the Amazon Review dataset.
To achieve ($\epsilon, \delta$)-differential privacy, where $\delta = 10^{-6}$ based on the training scale and $\epsilon = 2,4,8$, we set the noise scale $\sigma = 1.0, 0.77, 0.6$. 
As shown in Figure~\ref{fig:dp}, \name can still benefit FL training across different differential privacy guarantees.
 
\paragraph{Impact of malicious attacks.}
We investigate the robustness of \name by manually involving corrupted clients. 
Following a popular adversarial ML setting that introduces local model poisoning \citep{poison}, we randomly flip the ground-truth data labels for these corrupted clients. 
As shown in Figure~\ref{fig:malicious}, we introduce different percentages of corrupted clients to the OpenImg task.
We set the percentage of corruption below 15\%, which is a practical percentage under the real-world setting~\citep{poison}.
We observe that \name still improves performance across different degrees of corruption through identifying malicious clients and eliminating their interference.

\paragraph{Impact of unstable client.}
Finally, we show \name is robust with unstable clients who fail to maintain their affinity records, which may result in less accurate clustering results. 
We consider loss rates from 0\% to 20\% and report the corresponding final test accuracy in Figure~\ref{fig:client-fail}. 
We notice \name outperforms the baseline across different affinity loss rates.

%% file: pages/related.tex
 \section{Related Work}
  \label{sec:related}

\paragraph{Distributed Machine Learning}

Distributed ML in data centers has been well-studied~\citep{dml1,dml2,dml3}, where homogeneous data and workers are assumed~\citep{tiresias}.
With the same training goal, FL raises its unique challenges including the data heterogeneity and system heterogeneity.
As a result, \name aims at speeding up the training process through directly reducing the intra-cohort heterogeneity at scale.

\paragraph{Federated Learning}
FL is a distributed machine learning paradigm~\citep{tff-paper,flaas} with two key challenges: statistical and system heterogeneity.
State-of-the-art FL algorithms try to tackle these two challenges and optimize different targets including model convergence~\citep{yogi,prox,oort,hamed,liu2021projected,mf}, fairness~\citep{ditto,qfedavg}, privacy~\citep{privacy-pre,mycelium,honeycrisp,orchard}, efficiency~\citep{communication-eff,wenjun,compression,xie2023federatedscope}, and robustness~\citep{ditto,siren}.
However, they underperform in FL because they do not tackle the root cause of FL challenges but mitigate the negative effect caused by heterogeneity.  

\paragraph{Federated Analytics}
There has been significant work on geo-distributed data analytics~\citep{gaia-cmu, relay-hotcloud, CLARINET,Iridium}.
 They mainly optimize the execution latency~\citep{sol-nsdi} and resource efficiency~\citep{spanstore,fedchain}.
To further preserve privacy for distributed data, Orchard~\citep{orchard} and Honeycrisp~\citep{honeycrisp} have been proposed to enable large-scale differentially private analytics. 
Helen~\citep{helen} and Cerebro~\citep{cerebro} allow multiple parties to securely train models without revealing their data.

\paragraph{Traditional Clustering Algorithms}
Clustering algorithms~\citep{cluster-survey1,cluster-survey2} are used in popular data mining techniques, which usually assume access to all data.
However, under FL setting, it is non-trivial to design a clustering algorithm because of the unavailability of data.
\name proposes a clustering algorithm that can be applicable to the FL settings.

\paragraph{FL Client Clustering}
 In order to leverage the nature of clusters in real-world FL dataset, many algorithms have been proposed to identify the clusters among FL clients.
 However, existing clustered FL solutions~\citep{cfl,ifca,hc+fl,flexcfl}  mainly suffer from scalability and practicality, which are hard to adapt to large-scale, low-participation, and resource-constraint FL training.
Considering all real-world constraints, \name build a practical system to identify cohorts and benefit FL training.

%% file: pages/conclusion.tex
\section{Conclusion}
\label{sec:con}

We presented \name, which builds on top of the observation that there exist natural groups of statistically similar clients (cohorts) in large real-world FL populations. 
\name identifies cohorts with reduced intra-cohort heterogeneity at scale, addressing heterogeneity-borne FL challenges at their roots.
\name proposes an efficient algorithm and practical system that can be applied under real-world FL constraints to significantly benefit FL training in terms of model convergence, final accuracy, and model bias.

%% file: pages/ack.tex
\section*{ACKNOWLEDGMENTS} 

We are grateful to the anonymous reviewers, our shepherd Matthias Boehm and SymbioticLab members for their valuable comments and suggestions that improved the paper.
We thank the CloudLab team for providing GPU servers for \name experiments.  
This work was supported in part by NSF grant CNS-2106184 and a grant from Cisco.

%% file: pages/appendix.tex
 \section{Proof of Lemma~\ref{lm:res}}
 \label{sec:proof}

  We first make precise some definitions that are related to the proof from SCAFFOLD and then see the proof of Lemma~\ref{lm:res}. 
    
 \subsection{Additional definitions}
 
 \paragraph{Assumption 1.}
 $g_i(w)$ is unbiased stochastic gradient of $f_i$ with bounded variance, where $f_i$ represents the loss function on client $i$.
 
 $$
\mathbb{E}_{x_i} [ ||	g_i (w) - \nabla f_i (w)	||^2  ] \leq \sigma^2, \forall i,x.
 $$

 where $w$ is the aggregated server model.
  Note that $\sigma$ only bounds the variance within clients not across clients.
   \paragraph{Assumption 2.}
$\{ f_i \} $ are $\beta$-smooth and satisfy:
$$
||\nabla f_i(w) - \nabla f_i(v)|| \leq \beta ||w-v||, \forall i, w, v.
$$

      \paragraph{Assumption 3.}
	$f_i$ is $\mu$-convex for $\mu\geq0 $ and satisfies:
   $$
   	\langle  \nabla f_i(w) , v-w\rangle \leq -(\nabla f_i(w) - \nabla f_i(v) +\frac{\mu}{2}||w-v||^2), \forall i, w, v.
   $$

   \paragraph{Assumption 4.} 
(G, B)-BGD or Bounded Gradient Similarity: there exist constants $G\geq0$ and $B\geq1$ such that
 $$
 \frac{1}{N}\sum_{i=1}^{N}|| \nabla f_i(w)||^2 \leq G^2+B^2 \nabla f(w) ), \forall w.
 $$
   
 \subsection{Theoretical Results}

 \paragraph{ Lemma 1.}
 If the population and training resources are partitioned into up to $K$ cohorts, to theoretically achieve better model convergence, intra-cohort heterogeneity should  be reduced by $\sqrt{K}$ times 
when the training resource $|\mathcal{P}|$ is larger than $\alpha \sqrt{ \frac{  |\mathcal{P}_0| }{J_0^2}}$.
  $\alpha$ is a constant setting specified in SCAFFOLD that elaborates the relationship between model convergence and training resources.

 \paragraph{ \textit{Proof.} }
    
   We first borrow the proof of convergence analysis on FedAvg (Theorem 1) from SCAFFOLD following the same assumptions mentioned above: 

$$
\mathbb{E}[ f(w^R)] - f(w^*) \leq 3||w^0-w^*||^2\mu e^{-\frac{\widetilde{\eta}}{2} R } 
$$
\vspace{-0.4cm}
$$
 + \widetilde{\eta} (\frac{2\sigma^2}{kP} (1+\frac{P}{\eta^2_g}) +\frac{8G^2}{P}(1-\frac{P}{N})  )
 +  \widetilde{\eta}^2(36\beta G^2), 
$$
\vspace{-0.2cm} 
$$
\forall \frac{1}{\mu R}  \leq \widetilde{\eta} \leq \frac{1}{8(1+B^2)\beta}
$$

where P denotes the training resources, 
k is the number local steps, 
 $\eta_l$ is the local step-size, 
 $\eta_g$ is the global step-size 
 and $\widetilde{\eta} = k \eta_l \eta_g$ is the effective step-size

 Since we only care about the effect of training resources $P$ and heterogeneity $G$ on the convergence analysis, we further simplify the right hand side equation to be 
 $$
 h(P, G) =  \frac{\theta}{P} + \gamma \frac{G^2}{P} + \rho G^2 +\xi 
 $$
 where $ \theta, \gamma, \rho$ and  $\xi$ are constant settings.  
   Since we proportionally partition the population and training resources, we can assume $ (1-\frac{S}{N})$  to be constant before and after partition.
   
 In order to have no worse model convergence bound after partitioning, we need $h(P, G)$ to be non-increasing with the reduction of training resources $P$.
  As proposed in Lemma~\ref{lm:res}, \name partitions $K$ cohorts when the intra-cohort heterogeneity can be reduced by $\sqrt{K}$ times, which approximately give  $\frac{G^2}{P}$ be constant as the one before partition $\frac{G_0^2}{P_0}$.
 By substituting this relationship into $h(P, G)$, we can derive the lower bound for the range of training resources required to achieve better convergence bound:
 $$
 P \geq \sqrt{ \frac{\theta P_0  }{G_0^2\rho}} 
 = \sqrt{\frac{\sigma^2}{18 k  \widetilde{\eta}^2\beta}   \frac{P_0}{G_0^2}} = \alpha
 $$